%% file: icml26.tex
\documentclass{article}
\usepackage[accepted]{icml2026}
\usepackage{microtype}
\usepackage{graphicx}

\usepackage{hyperref}

\usepackage{url}
\usepackage{algorithmic,algorithm}
\usepackage{booktabs, multirow}
\usepackage{caption}
\usepackage{wrapfig}
\usepackage{enumitem}

\definecolor{mygray}{gray}{0.6}
\usepackage{colortbl}
\usepackage{tcolorbox}
\usepackage{hyperref}
\usepackage{subcaption}
\usepackage{amsmath}
\usepackage{array}
\usepackage{ragged2e}

\usepackage[capitalise,nameinlink]{cleveref}
\usepackage[dvipsnames]{xcolor}
\newcommand{\mscellf}{\rowcolor{teal!8}}
\newcommand{\mscells}{\rowcolor{teal!15}}
\newcommand{\mscellt}{\rowcolor{teal!20}}
\newcommand{\mscellfo}{\rowcolor{teal!28}}

\definecolor{custom_blue}{HTML}{044dc1}  

\hypersetup{
    colorlinks=true,
    linkcolor=custom_blue,
    citecolor=custom_blue,
    filecolor=cyan,
    urlcolor=purple
}

\usepackage{amsmath}
\usepackage{amssymb}
\usepackage{mathtools}
\usepackage{amsthm}

\theoremstyle{plain}

\theoremstyle{definition}

\theoremstyle{remark}

\usepackage[textsize=tiny]{todonotes}

\icmltitlerunning{EpiCache: Episodic KV Cache Management for Long-Term Conversation on Resource-Constrained Environments}
\newcommand{\methodname}{{\textsc{EpiCache}}} 

\usepackage[textsize=tiny]{todonotes}

\begin{document}

\twocolumn[
  \icmltitle{EpiCache: Episodic KV Cache Management for Long-Term \\Conversation on Resource-Constrained Environments}



  \icmlsetsymbol{equal}{*}

  \begin{icmlauthorlist}
    \icmlauthor{Minsoo Kim}{comp}
    \icmlauthor{Arnav Kundu}{comp}
    \icmlauthor{Han-Byul Kim}{comp}
    \icmlauthor{Richa Dixit}{comp}
    \icmlauthor{Minsik Cho}{comp}
  \end{icmlauthorlist}

  \icmlaffiliation{comp}{Apple}
  
  \icmlcorrespondingauthor{Minsoo Kim}{minsoo@apple.com}
  \icmlcorrespondingauthor{Minsik Cho}{minsik@apple.com}

  \icmlkeywords{Machine Learning, ICML}

  \vskip 0.3in
]



\printAffiliationsAndNotice{}  

\begin{abstract}
  Modern large language models (LLMs) extend context lengths to millions of tokens, enabling coherent, personalized responses grounded in long conversational history. However, the Key-Value (KV) cache grows linearly with the extended dialogue history, causing the model’s memory footprint to quickly exceed device limits. While recent KV cache compression methods attempt to reduce memory usage, most apply cache eviction after processing the entire context, incurring unbounded peak memory usage. Additionally, query-dependent eviction narrows the cache semantics to a single query, leading to failure cases in multi-turn conversations. In this paper, we introduce \methodname, a training-free KV cache management framework for long conversational question answering (LongConvQA) under fixed memory budgets. \methodname\ bounds cache growth through block-wise prefill and preserves topic-relevant context via episodic KV compression, which clusters conversation history into coherent episodes and performs episode-specific KV cache eviction. Across three LongConvQA benchmarks (LongMemEval, Realtalk, and LoCoMo), \methodname\ improves accuracy by up to 30\%, achieves near-full-cache accuracy under $4$–$6\times$ compression, and reduces latency and peak memory by up to $2.4\times$ and $3.7\times$, respectively. 
  
\end{abstract}
\input{Section/Intro}
\input{Section/Background}
\input{Section/method}
\input{Section/Experiment}

\input{Section/Related}
\input{Section/conclusion}

\bibliography{example_paper}
\bibliographystyle{icml2026}

\input{Section/Appendix}

\end{document}

%% file: Section/Intro.tex
\section{Introduction}

Large language models (LLMs)~\citep{gpt3,yang2025qwen3,touvron2023llamaopenefficientfoundation,jiang2023mistral7b} have significantly extended their maximum context lengths, with LLM-based AI assistants now capable of processing millions of tokens~\citep{reid2024gemini,llama4}. This capability enables assistants to leverage extensive dialogue histories when generating responses, producing personalized and contextually coherent outputs~\citep{openai2024gpt4technicalreport,claude3anthropic}, which are central requirements for conversational AI applications~\citep{FU202214}.
  
To benchmark whether assistants can leverage long conversational histories, recent work has formalized Long Conversational Question Answering (LongConvQA) using both human-human conversations and user-AI assistant interactions~\citep{maharana-etal-2024-evaluating,lee2025realtalk,wu2025longmemeval}. In this setting, an assistant must answer a sequence of user questions grounded in dialogue histories spanning multiple sessions over days or weeks. Most existing systems tackle LongConvQA by maintaining a retrieval-based memory bank that incrementally summarizes and stores conversation content and retrieves relevant entries for each query~\citep{Zhong_Guo_Gao_Ye_Wang_2024,mem0,pan2025secom,tan-etal-2025-prospect}. However, such systems rely on repeated API-based LLM calls to maintain the memory bank and do not bound memory usage  at inference time, limiting their applicability to resource-constrained environments.

We study how to enable LongConvQA in resource-constrained environments by designing a Key-Value (KV) cache management framework. The KV cache stores the Key and Value states of each token for reuse in auto-regressive generation, but its size grows linearly with context length, creating severe challenges in extended conversations. For instance, in multi-day dialogues between a user and an assistant~\citep{wu2025longmemeval}, the KV cache of LLaMA3.2-3B exceeds 7GB after only 30 sessions—larger than the size of the model parameters. This underscores the importance of cache management for deploying conversational AI systems on resource-constrained devices such as smartphones, where memory usage is strictly limited~\citep{alizadeh-etal-2024-llm, liu2024mobilellmoptimizingsubbillionparameter}.

Prior work has attempted to mitigate the growing memory usage of KV cache through various compression techniques~\citep{zhang2023ho,li2024snapkv,cai2025pyramidkvdynamickvcache}. Yet two major limitations remain in achieving LongConvQA under resource-constrained environments. First, most methods apply compression after prefill of the entire input context (\textit{post-prefill}), causing peak memory usage that scales linearly with input length. Second, \textit{query-dependent} eviction method~\citep{li2024snapkv} retains cache entries by focusing on the current query and may evict evidence required in later turns, degrading accuracy in multi-turn conversations.

To address these challenges, we propose \methodname, a training-free KV cache management framework that enforces a constant memory footprint through \textbf{block-wise prefill} and \textbf{episodic organization}. After processing each block, we evict less critical KV entries to free space for the next block, ensuring memory consumption is bounded. Based on block prefill, \methodname\ incorporates an episodic clustering method inspired by conversation segmentation studies~\citep{Joty_2013,galley-etal-2003-discourse}. Specifically, we apply clustering to group conversation history into coherent episodes, and perform episodic KV cache compression, yielding topic-specific caches while using constrained memory.

Finally, we find that LLMs exhibit different sensitivities to block prefill eviction across layers. Building on this observation, we propose an adaptive layer-wise budget allocation strategy that distributes the KV cache budget proportionally to each layer's sensitivity. Together with episodic KV cache management, this enables \methodname\ to preserve long-term conversation history while operating under strict memory limits, yielding up to 30\% higher LongConvQA accuracy than recent baselines and sustaining accuracy comparable to full KV under $4$–$6\times$ cache compression. In addition, our block-wise cache control framework reduces peak memory usage by up to $3.5\times$, while cache eviction accelerates decoding, cutting decoding latency by up to $2.4\times$ compared to full KV.

%% file: Section/Background.tex
\input{figs/fig1/fig1}
\input{figs/fig2/fig2}
\section{Background}\label{sec:background}
We begin by formalizing Long Conversational Question-Answering (LongConvQA) in \Cref{sec:longconvqa}. We then discuss memory-constrained KV cache management in \Cref{sec:mc-kv}, comparing post- and block-prefill eviction and discussing the resulting accuracy-memory trade-offs. Finally, in \Cref{sec:compression}, we review attention-guided cache compression with patched prompts and present analyses that motivate our method, \methodname.
\subsection{Long Conversational QA Formulation}
\label{sec:longconvqa}
We formalize LongConvQA as answering a sequence of user queries $\mathcal{Q} = \{q_1, \dots, q_{N_q}\}$, with $N_q$ denoting the total number of queries, conditioned on long conversational history~\citep{maharana-etal-2024-evaluating,lee2025realtalk,wu2025longmemeval}. Let the dialogue history be represented as an ordered sequence of $N_u$ utterances where each utterance $u_j$ pairs a role $r_j$ with text $t_j$: 
\begin{equation}
\begin{aligned}
\mathcal{H} &= \{u_1, u_2, \dots, u_{N_u}\}, \\
u_j &= (r_j, t_j), \quad 
r_j \in \{\text{speaker}_1, \text{speaker}_2\}
\end{aligned}
\label{eq:utter}
\end{equation}
Given a long conversation $\mathcal{H}$, an LLM encodes it into a Key-Value (KV) cache $\mathrm{KV}_{\mathcal{H}}$. For $L$ layers and $H$ KV heads, encoding $N$ tokens produces $L \times H \times N$ KV entries, growing linearly with the conversation length. In this work, we focus on token-level cache compression, where KV entries of less important tokens are evicted; the resulting compressed cache is denoted as $\widetilde{\mathrm{KV}}_{\mathcal{H}} \subseteq \mathrm{KV}_{\mathcal{H}}$, and we use the terms compression and eviction interchangeably.

Our goal is to generate accurate answers for all queries ${q_1, \dots, q_{N_q}}$ grounded in the dialogue history $\mathcal{H}$, using a compressed cache $\widetilde{\mathrm{KV}}_{\mathcal{H}}$ that satisfies a memory budget $M$, i.e., with size $L \times H \times M$, while preserving answers comparable to full KV cache ($\mathrm{KV}_{\mathcal{H}}$) based generation:
\begin{equation}
f_{\text{LM}}(q_i \mid \widetilde{\mathrm{KV}}_{\mathcal{H}}) \approx 
f_{\text{LM}}(q_i \mid \mathrm{KV}_{\mathcal{H}}),
\quad i = 1, \dots, N_q.
\label{eq:convqa}
\end{equation}
This formulation evaluates performance in a \textit{multi-turn conversational} setting, where multiple query–answer pairs are grounded in a shared dialogue history.

\subsection{KV Cache Management: Post vs. Block Prefill}
\label{sec:mc-kv}
Most existing KV compression approaches reduce cache size in the decoding stage by performing eviction \textit{after} the full context has been prefilled, i.e., \textit{Post Prefill Eviction}~\citep{li2024snapkv,feng2024ada,cai2025pyramidkvdynamickvcache,kim2025kvzipqueryagnostickvcache}. As shown in \Cref{fig:prelim_a}, this design causes peak memory usage to grow linearly with input length, since the entire context must be cached before any eviction takes place. Even with optimized attention kernels~\citep{dao2023flashattention2}, memory demandthe prefill stage remains unbounded, as observed in \Cref{fig:prelim_c} top.

To bound memory growth, \textit{Block Prefill Eviction}~\citep{kim-etal-2024-infinipot,corallo-papotti-2024-finch,park2025keydiffkeysimilaritybasedkv} processes the input in a block-wise way, handling one segment at a time under a fixed budget $M$. Each step adds $M_{\text{block}}$ tokens, after which an eviction step reduces KV cache entries back to $M$. For example, in \Cref{fig:prelim_b}, the budget is $M=1$, and each block adds $M_{\text{block}}=3$ tokens that are then evicted back to $M$. This design ensures that the number of cache entries never exceeds $M+M_{\text{block}}$, keeping peak GPU memory usage flat with input length as highlighted in \Cref{fig:prelim_c} top.

However, this bounded memory comes with a steep accuracy trade-off: when the state-of-the-art post prefill eviction method KVzip,~\citep{kim2025kvzipqueryagnostickvcache,kvpress_leaderboard} is applied in the block prefill setting, LongConvQA scores~\citep{maharana-etal-2024-evaluating} degrade sharply across all budget levels. This underscores a central challenge—while block prefill guarantees constant memory usage, adapting post-prefill eviction methods to this setting severely undermines answer quality in LongConvQA. 

\input{figs/fig3/fig3}

\subsection{Attention-guided KV Cache Compression}
\label{sec:compression}

To address the accuracy degradation of block prefill eviction, prior work employs attention-based token scoring with a patched prompt. Here, token importance is quantified by the cross-attention it receives from query tokens: $\mathrm{Attn}(x_t \!\rightarrow\! x_i)$ which denotes the attention weight from a query token $x_t$ to a key token $x_i$. Tokens that receive higher attention from queries are considered more important, while those with lower scores are evicted to satisfy the memory budget $M$.

To provide guidance for cache eviction, the \textit{patched prompt} strategy~\citep{kim-etal-2024-infinipot, corallo-papotti-2024-finch} appends an auxiliary prompt of length $p$ after each block ending at token $n$. These queries ${x_{n+1},\dots,x_{n+p}}$ attend back to the preceding block tokens ${x_{1},\dots,x_{n}}$, as illustrated in \Cref{fig:prelim_b}. The resulting importance score $s_i$ of token $i$ is aggregated by taking the maximum across token-axis~\citep{kim2025kvzipqueryagnostickvcache}, as defined in \Cref{eq:patched_prompt}. The patched prompt is only used for scoring and not retained in the KV cache.
\begin{equation}
\begin{aligned}
s_i^{\text{max}} 
&= \max_{t \in [n+1,n+p]} \mathrm{Attn}(x_t \!\rightarrow\! x_i).
\end{aligned}
\label{eq:patched_prompt}
\end{equation}

In block prefill, accuracy strongly depends on the content of the patched prompt. 
To quantify this effect, \Cref{fig:distance} presents a controlled experiment that assumes oracle access to the future user query and inserts it directly as the patched prompt (Exact-Question), yielding the highest accuracy under a fixed memory budget. 
While this setting provides an upper bound on performance, it is infeasible in LongConvQA for two reasons~\citep{kim2025kvzipqueryagnostickvcache}: 
(i) future user queries are unknown at compression time, and 
(ii) such \textit{query-dependent} compression requires re-prefilling the entire dialogue history for every new question.

Since the dialogue history $\mathcal{H}$ consists of query-answer turns, it offers an opportunity to approximate the future query with semantically related turns: we embed the future queries and all utterances $u_1,\dots,u_{N_u}$ into a shared embedding space, rank utterances by semantic similarity, and construct the patched prompt from the top-ranked ones. Using the resulting patched-prompt, we run block prefill eviction and evaluate LongConvQA accuracy. As shown in \Cref{fig:distance}, the prompts formed from the most semantically aligned utterances (top 10\%) nearly match Exact-Question performance, while accuracy degrades as semantic alignment decreases. This observation motivates our goal: identifying, without access to future queries, patched prompts that approximate unseen questions. To this end, we introduce an unsupervised clustering method to discover dialogue segments aligned with future queries (\Cref{sec:episodic}).

%% file: figs/fig1/fig1.tex
\begin{figure*}[ht]
    \centering
    \begin{subfigure}[t]{0.24\textwidth}
    \captionsetup{justification=centering}
        \includegraphics[width=\textwidth]{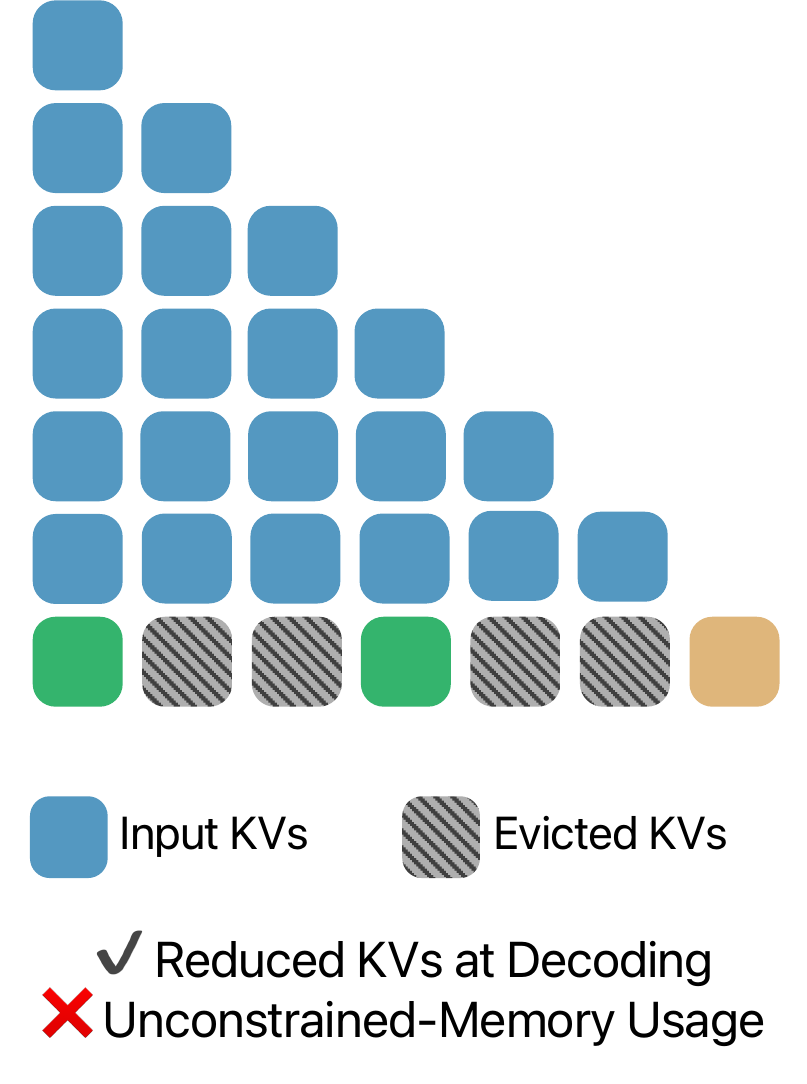}
        \subcaption{Post Prefill Eviction}
        \label{fig:prelim_a}
    \end{subfigure}
    \centering
    \begin{subfigure}[t]{0.32\textwidth}
    \captionsetup{justification=centering}
        \includegraphics[width=\textwidth]{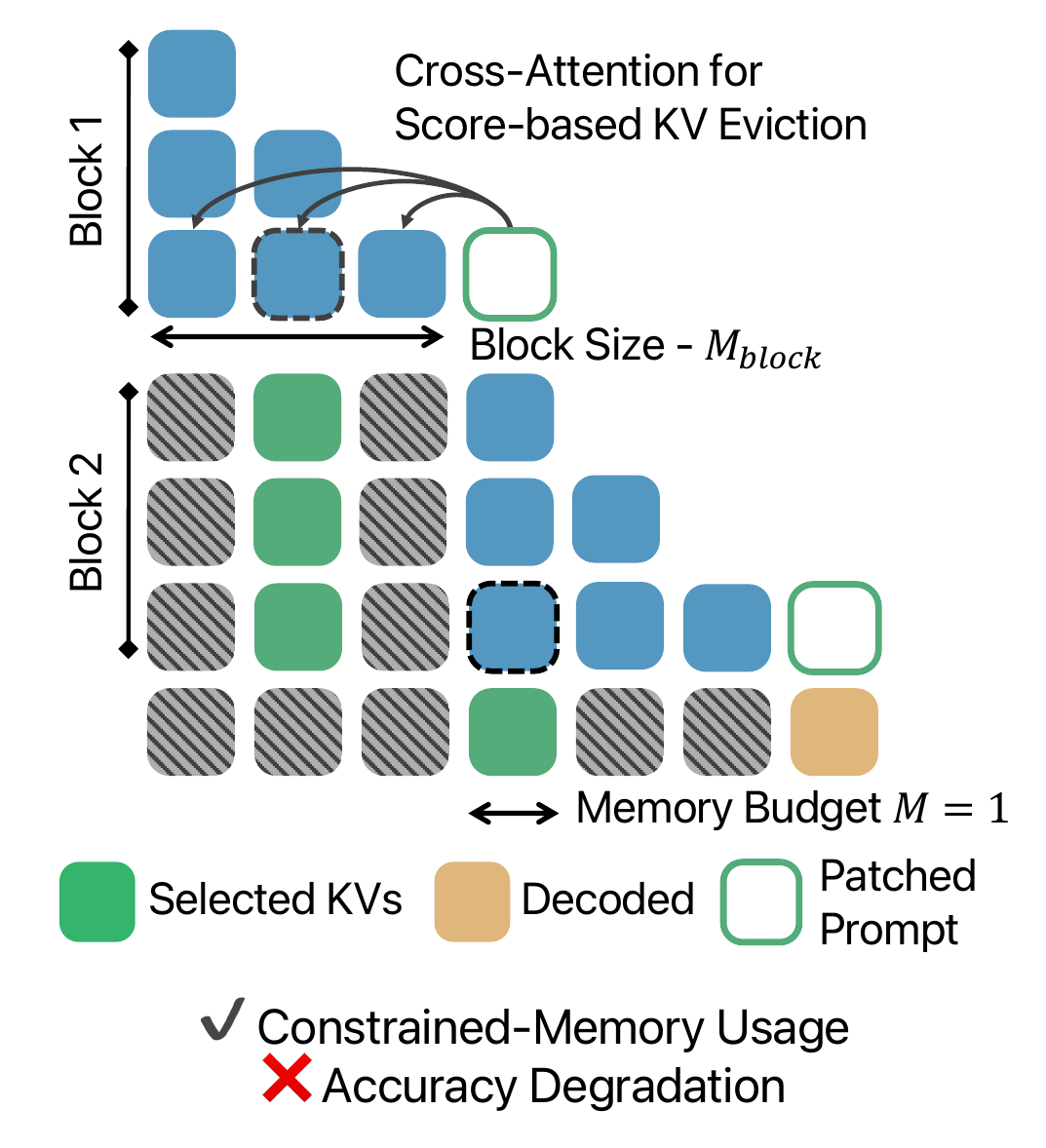}
        \subcaption{Block Prefill Eviction}
        \label{fig:prelim_b}
    \end{subfigure}
    \centering
    \begin{subfigure}[t]{0.27\textwidth}
    \captionsetup{justification=centering}
        \includegraphics[width=\textwidth]{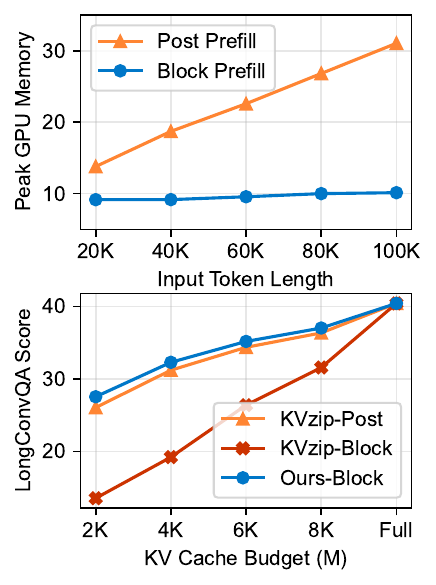}
        \subcaption{Top: Peak GPU Memory Bottom: LongConvQA Score}
        \label{fig:prelim_c}
    \end{subfigure}
    \caption{\textbf{KV Cache Management Analysis.} (a) Post prefill eviction: eviction after full-context prefill, reducing KV size at decoding but causing unbounded memory usage. (b) Block prefill eviction: input processed in 3-token blocks with patched prompts for scoring, then evicted down to 1 token. (c) Top: Peak GPU memory vs. input length on LLaMA-3.2-3B with A100. Bottom: LongConvQA accuracy of KV compression methods under post vs. block prefill on LLaMA-3.2-3B.}
    \label{fig:prelim}
    \vspace{-.15in}
\end{figure*}

%% file: figs/fig2/fig2.tex
\begin{figure}[ht]
\centerline{\includegraphics[width=0.85\columnwidth]{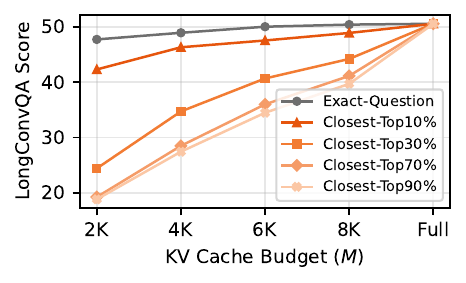}}
\caption{\textbf{Patched-Prompt Analysis}: LoCoMo results with LLaMA3.1-8B under block prefill. Patched prompts are formed by selecting the top 10\%–90\% similar conversation utterances to $q_i$.}\label{fig:distance}
\end{figure}

%% file: figs/fig3/fig3.tex
\begin{figure*}[ht]
    \centering
    \begin{subfigure}[t]{0.273\textwidth}
    \captionsetup{justification=centering}
        \includegraphics[width=\textwidth]{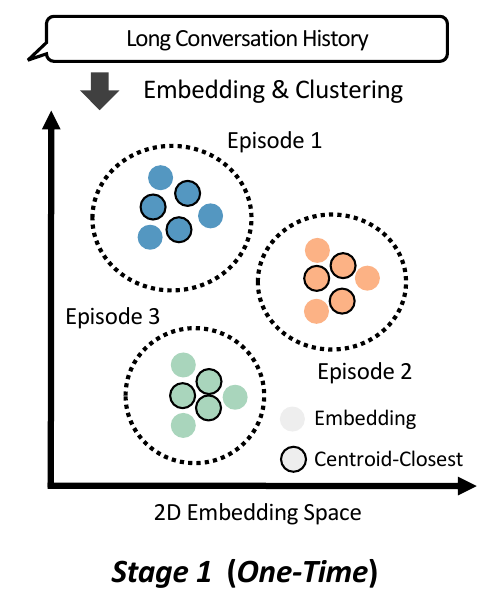}
        \subcaption{Conversation Clustering}
        \label{fig:main_method_a}
    \end{subfigure}
    \centering
    \begin{subfigure}[t]{0.27\textwidth}
    \captionsetup{justification=centering}
        \includegraphics[width=\textwidth]{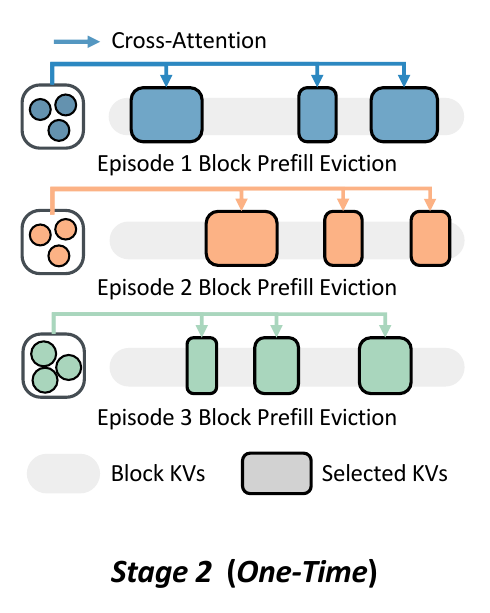}
        \subcaption{Building Episodic KV Cache}
        \label{fig:main_method_b}
    \end{subfigure}
    \centering
    \begin{subfigure}[t]{0.27\textwidth}
    \captionsetup{justification=centering}
        \includegraphics[width=\textwidth]{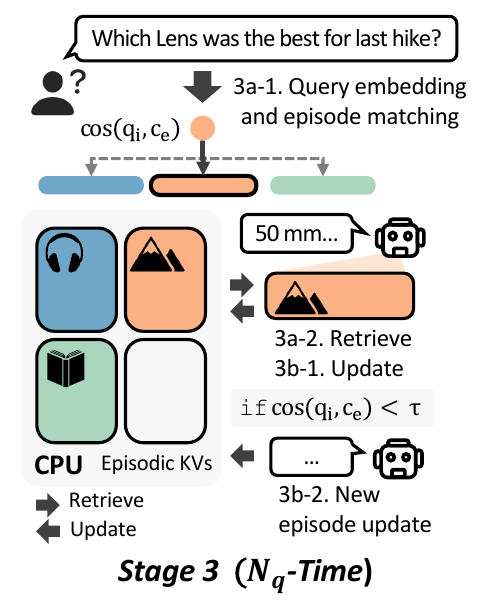}
        \subcaption{Decoding with Episodic KVs}
        \label{fig:main_method_c}
    \end{subfigure}
    \caption{
    \textbf{\methodname ~Overview.} (a) segmentation and embedding of the conversation, followed by clustering into episodes. (b) building episodic KV caches under a fixed memory usage based on representative segments of each episode. (c) an incoming query is embedded, matched to the closest episode, and the corresponding cache is retrieved for answer generation, and updated with newly generated tokens.}
    \label{fig:main_method}
    \vspace{-.15in}
\end{figure*}

%% file: Section/Method.tex
\section{Method}
\label{sec:method}

\subsection{Episodic KV Cache Management } 
\label{sec:episodic}
Conversations span multiple topics within a single session~\citep{maharana-etal-2024-evaluating,pan2025secom}. Such long histories can be segmented into coherent episodes, where subsequent utterances are grounded in different episodes of the prior dialogue. This motivates \methodname: by clustering long-term conversation histories into multiple episodes and constructing episode-specific caches, we match each incoming query to the most relevant cache for accurate answer generation. As illustrated in \Cref{fig:main_method}, this process unfolds in three stages:  offline conversation clustering, episodic KV cache construction, and query-matched decoding.


\paragraph{Stage 1. Conversation Clustering.~(\Cref{fig:main_method_a})}
For clustering conversation, we first divide the raw dialogue history $\mathcal{H}$ into segments of $w_{\text{embed}}$ utterances, denoted as $\mathcal{H} = {S_1, \dots, S_K}$.
\begin{equation}
\begin{aligned}
S_k &= {u_{(k-1)w_{\text{embed}}+1}, \dots, u_{\min(kw_{\text{embed}},N_u)}}, \\
k &= 1,\dots,K,\quad 
K = \left\lceil \tfrac{N_u}{w_{\text{embed}}} \right\rceil
\end{aligned}
\end{equation} 
Each segment $S_k$ is encoded with a lightweight encoders~\citep{qwen3embedding} $f_{\text{embed}}$ into a vector embedding $\mathbf{e}_k \in \mathbb{R}^d$, capturing the segment’s semantics. We then apply K-Means clustering $\mathcal{C}(\cdot)$ to the embeddings $\{\mathbf{e}_k\}_{k=1}^K$:
\begin{equation}
\mathcal{C}(\{\mathbf{e}_k\}_{k=1}^K) \to 
\{\mathcal{E}_1,\dots,\mathcal{E}_E\}, 
\ 
\bigcup_{e=1}^E \mathcal{E}_e=\{S_1,\dots,S_K\}.
\label{eq:clustering}
\end{equation}
\Cref{eq:clustering} partitions $\mathcal{H}$ into $E$ semantically coherent topical episodes~\citep{clustering_dialogue}.\footnote{Qualitative analysis of clustering is provided in Appendix~\ref{appn:analysis_clustering}.}
For each cluster \(\mathcal{E}_e\), we can compute its centroid embedding $c_e$: 
\begin{equation} \mathbf{c}_e = \frac{1}{|\mathcal{E}_e|} \sum_{S_k \in \mathcal{E}_e} \mathbf{e}_k, \quad S_{\text{centroid-closest}}^{(e)} = \arg\max_{S_k \in \mathcal{E}_e} \cos(\mathbf{e}_k, \mathbf{c}_e). 
\label{eq:centroid}\end{equation}

We then identify the \textit{representative} segments of the cluster $S_{\text{centroid-closest}}^{(e)}$—i.e., the conversation segment in each cluster whose embedding is closest to the centroid in terms of cosine similarity. These centroid-closest segments contain multiple turns from both speakers and is used as the patched prompt in the subsequent block prefill eviction step. 

\paragraph{Stage 2. Episodic KV Cache Construction.~(\Cref{fig:main_method_b})}
As discussed in \Cref{sec:compression}, patched prompts guide cache eviction toward retaining tokens relevant to the prompt content. Building on this insight, \methodname\ uses the centroid-closest segments of each episode as the patched prompt, thereby collecting episode-specific KV entries under the memory budget $M$. 

For each episode $e \in \{1,\dots,E\}$, we perform block prefill eviction over the entire context, appending its patched prompt after each block. Attention scores are then computed as in~\cref{eq:patched_prompt}, and the top $M$ tokens are retained to form an episode-specific cache $C_{\mathrm{KV}}^{(e)}$. Finally, all episodic caches are collected into $\mathbb{B} = \{ C_{\mathrm{KV}}^{(1)}, \dots, C_{\mathrm{KV}}^{(E)} \}$ and stored offline for later retrieval.


\paragraph{Stage 3a. KVs Matching and Decoding.~(\Cref{fig:main_method_c})}
At decoding time, each user query $q_i$ is embedded with the same encoder $f_{\text{embed}}$ used in clustering, ensuring that it lies in the same representation space as the episode centroids. The query is then matched to the closest centroid as follows:
\begin{equation}
\mathbf{q}_i = f_{\text{embed}}(q_i),
\quad
e^\dagger = \arg\max_{e \in [1,E]} \cos(\mathbf{q}_i, \mathbf{c}_e).
\end{equation}
As illustrated in ~(3a-2) of \Cref{fig:main_method_c}, the framework retrieves the corresponding episodic KV cache $C_{\mathrm{KV}}^{(e^\dagger)}$ from $\mathbb{B}$ and conditions generation on it: $f_{\text{LM}}(q_i \mid C_{\mathrm{KV}}^{(e^\dagger)},)$. This design enables query-specific retrieval while keeping cache size bounded under memory budget $M$.

\paragraph{Stage 3b. Episodic KV Cache Update.~(\Cref{fig:main_method_c})}
For each turn, \methodname\ manages episodic KV caches in two ways to support a continuous conversational experience.

\begin{itemize}
    \vspace{-0.1in}\item \textbf{Update an existing episode as in (3b-1) of \Cref{fig:main_method_c}):} When an episodic KV cache is retrieved for decoding as in Stage~3a, the KV pairs generated at the current turn are appended to the retrieved episodic cache, enriching episode-specific memory. If a subsequent turn is matched to a different episode due to a topic shift, the episodic cache used in the previous turn—appended with newly generated KV pairs—is written back to the episodic KV cache set. 
    \item \textbf{Create a new episode as in (3b-2) of \Cref{fig:main_method_c}):} If the query embedding is not sufficiently similar to any existing centroid ($\max_e \cos(\mathbf{q}_i, \mathbf{c}_e) < \tau$), \methodname\ creates a new episode. A new episodic KV cache is initialized from the KV pairs of the new turn and added to the cache set $\mathbb{B}$, allowing previously unseen topics to be incorporated into the episodic structure. 
\end{itemize}

To maintain strict memory bounds, \methodname\ refreshes an episodic cache only when its size exceeds $M + M_{\text{block}}$. Upon reaching this cache budget, we collect the newly accumulated episode history $\mathcal{H}'$ and re-run the same clustering procedure as in Stage~1. We then update the representative segment for the episode, and rebuild the episodic caches by block prefill eviction guided by the updated prompt under the fixed budget $M$. An interesting extension is to update the compressed episodic KV cache using only newly added turns, which we leave as future work.

\subsection{Sensitivity-aware Layer-wise KV Budget Allocation}
\label{sec:adaptive}
We further address the accuracy degradation of block prefill by proposing a KV cache \textit{budget allocation} strategy. The key idea is to measure how much each layer's Key states representation deviates under block prefill and to distribute a KV budget across layers in proportion to this deviation.

\input{figs/fig4/fig4}

\paragraph{Simulating Block Prefill via Custom Masking.}
To quantify the deviation caused by block prefill eviction, we introduce a custom masking scheme. Each transformer layer is represented as a function $f$ that takes the previous layer's output $X^{(\ell-1)} \in \mathbb{R}^{N \times d}$. Here, $N$ denotes the input sequence length and $d$ the hidden dimension. The function produces the $\ell_{\text{th}}$ layer output $X^{\ell} = f\left(X^{\ell-1}, \mathcal{M}\right)$, where $\mathcal{M}$ is the standard causal mask.

We replace $\mathcal{M}$ with a custom mask $\mathcal{M}'$ that enforces a budget $M$, attending to sink tokens and the most recent tokens~\citep{xiao2024efficient}. This design follows static KV cache compression methods, allowing us to simulate block prefill eviction in a single forward pass and directly measure its effect on layer representations.

\input{figs/fig5/fig5}
\paragraph{Layer Sensitivity Guided KV Budget Allocation.}
We quantify the per-layer impact of block prefill eviction by comparing Key states\footnote{Further details regarding the rationale for using Key states deviation are provided in Appendix~\ref{appn:analysis_sensitivity}.}  produced under the causal mask $\mathcal{M}$ and the custom mask $\mathcal{M}'$. For each layer $\ell$, the forward pass under each mask produces:
\begin{equation}
\begin{aligned}
K^{\ell}_{\text{full}} &= f\left(X_{\text{full}}^{\ell-1}, \mathcal{M}\right) W_K^{\ell}, \\
K^{\ell}_{\text{block}} &= f\left(X_{\text{block}}^{\ell-1}, \mathcal{M}'\right) W_K^{\ell}
\end{aligned}
\label{eq:mask}
\end{equation}
where $K^{\ell}_{\text{full}}$ and $K^{\ell}_{\text{block}}$ are the $l$-th layer Key states computed under $\mathcal{M}$ and $\mathcal{M}'$, respectively. We then define layer sensitivity as the average cosine similarity between the two sets of Key vectors across attention heads and input tokens:
\begin{equation}
\sigma_\ell = \tfrac{1}{HN}\sum_{h=1}^H \sum_{i=1}^N 
\cos\!\big(k^{(\ell,h)}_{\text{full},i},\,k^{(\ell,h)}_{\text{block},i}\big)
\label{eq:mask_diff}
\end{equation}
Empirically, $\sigma_\ell$ exhibits large variation across layers yet remains consistent across different inputs in \Cref{fig:budget_alloc_a} (shadowed regions denote input variance), indicating that layer sensitivity is \textit{model-dependent} rather than input-dependent. We define $s_\ell = 1 - \sigma_\ell$ as the sensitivity score for layer $\ell$. Based on this observation, we propose a sensitivity-aware budget allocation strategy that assigns larger cache budgets to layers more sensitive to block prefill and smaller budgets to less sensitive ones. Specifically, we redistribute the global budget ($M\cdot L$) according to layer sensitivity scores $s_\ell$, with $\alpha$ controlling how sharply the allocation emphasizes sensitive layer:
\begin{equation}
M_\ell^{\text{alloc}} 
= \frac{s_\ell^{\,\alpha}}{\sum_{j=1}^{L} s_j^{\,\alpha}} \cdot (L \cdot M),
\quad 
\sum_{\ell=1}^{L} M_\ell^{\text{alloc}} = L \cdot M
\end{equation}\label{eq:alloc}
We evaluate the sensitivity-aware approach by measuring how budget allocation shifts the KL divergence between answer predictions generated from block prefill eviction and those generated from the full KV cache, where negative values indicate closer alignment with full KV answer generation. As shown in \Cref{fig:budget_alloc_b}, sensitivity-aware allocation shifts KL divergence by –0.80 relative to uniform allocation. In contrast, recent budget allocation methods such as PyramidKV~\citep{cai2025pyramidkvdynamickvcache}, which follows a pyramid-shaped budgeting, and retrieval head profiling based allocation~\citep{wu2025retrieval} tend to increase KL divergence. Our reduced KL divergence is reflected in improved LongConvQA accuracy (see \Cref{fig:main_exp}), as further discussed in \Cref{subsec:main_results}.


%% file: figs/fig4/fig4.tex


\begin{figure}[t]
    \centering
    \begin{subfigure}[t]{0.235\textwidth}
    \captionsetup{justification=centering}
        \includegraphics[width=\textwidth]{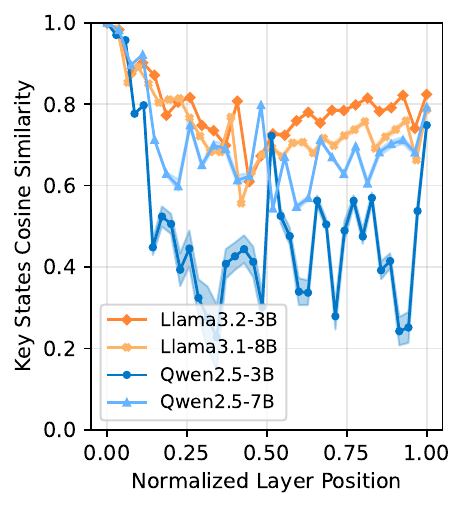}
        \subcaption{Key states cosine similarity}
        \label{fig:budget_alloc_a}
    \end{subfigure}
    \centering
    \begin{subfigure}[t]{0.241\textwidth}
    \captionsetup{justification=centering}
        \includegraphics[width=\textwidth]{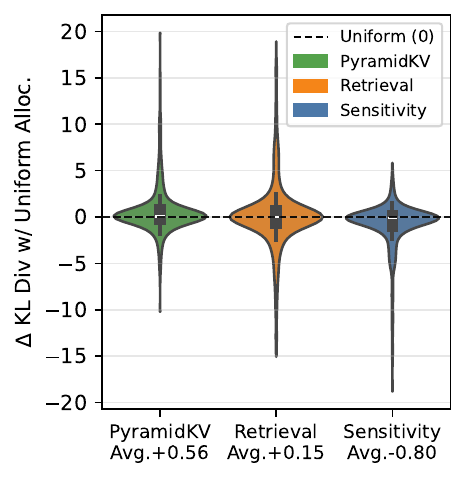}
        \subcaption{$\Delta$ KL divergence w/ full KV}
        \label{fig:budget_alloc_b}
    \end{subfigure}
    \centering
    \vspace{-.05in}
    \caption{\textbf{Layer-wise Sensitivity Analysis.} (a) Key states cosine similarity across layer positions. (b) KL divergence measured between block prefill ($M{=}4$K) and full KV answer predictions, with uniform allocation as the baseline. Qwen2.5-7B used.}
    \label{fig:budget_alloc}
    \vspace{-.2in}
\end{figure}

%% file: figs/fig5/fig5.tex
\begin{figure*}[ht]
    \centering
    \includegraphics[width=0.88\textwidth]{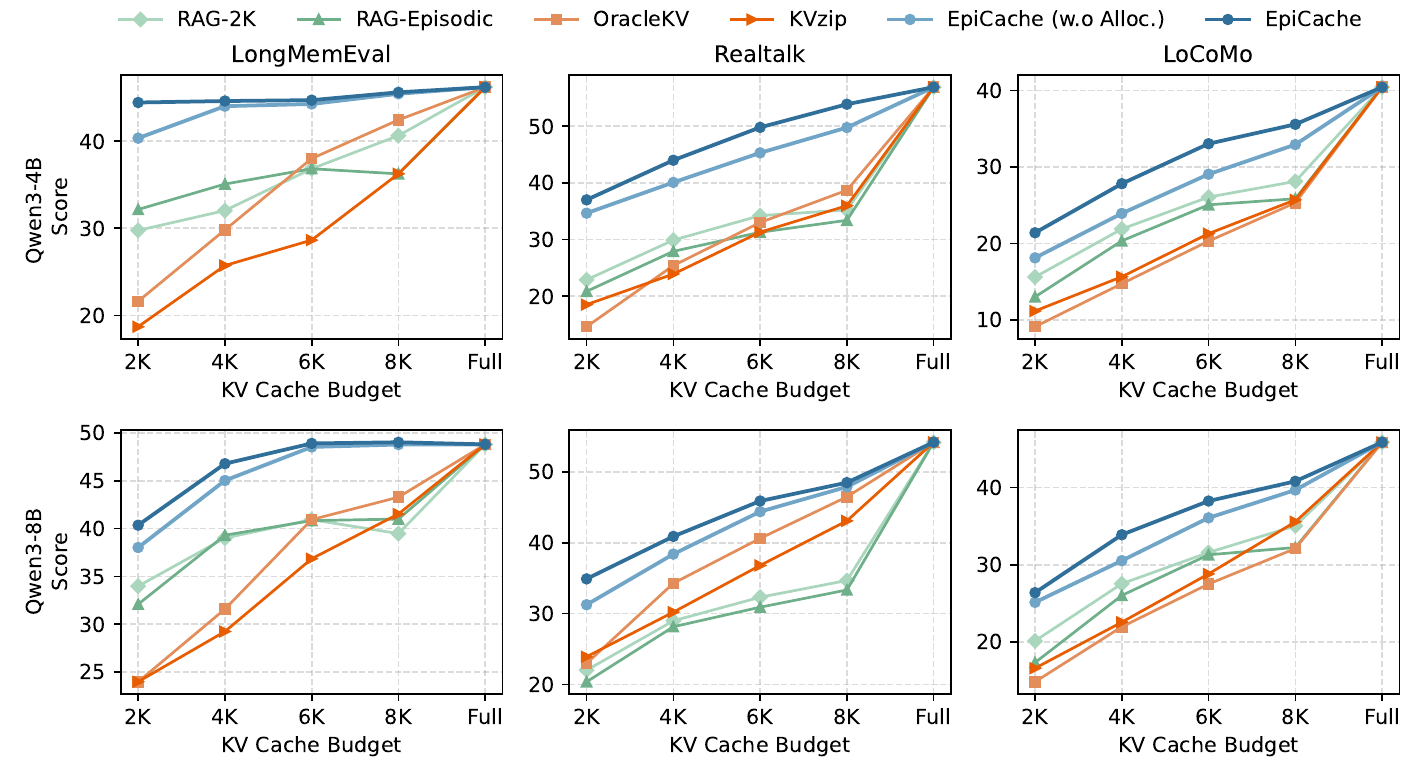}
    \vspace{-0.1in}
    \caption{\textbf{LongConvQA Evaluation} (LongMemEval, Realtalk, and LoCoMo) results with fixed KV cache budgets $M\in\{2\text{K},4\text{K},6\text{K},8\text{K}\}$ across Qwen3 family models. The number of episodes (clusters) fixed to $E{=}4$ in all experiments.}
    \label{fig:main_exp}
    \vspace{-0.2in}
\end{figure*}

%% file: Section/Experiment.tex
\section{Experiments}
\label{sec:exp}

\paragraph{Models and Benchmarks.}We evaluate on a comprehensive family of models: Qwen3~\citep{yang2025qwen3}, LLaMA3.1 and 3.2~\citep{grattafiori2024llama} and Qwen2.5~\citep{qwen2025qwen25technicalreport}. All evaluations follow the LongConvQA formulation in \Cref{eq:convqa}, where models answer various queries grounded in long conversation histories. 
We use three benchmarks: Realtalk~\citep{lee2025realtalk} and LoCoMo~\citep{maharana-etal-2024-evaluating}, containing multi-day human-human dialogues, and LongMemEval~\citep{wu2025longmemeval}, consisting of multi-session user-LLM dialogues. Further details of benchmarks are provided in \Cref{appn:dataset}.

\paragraph{Baselines.} (1) RAG: As an alternative memory-constrained design for LongConvQA, we consider retrieval-augmented generation~\cite{lewis2021retrievalaugmentedgenerationknowledgeintensivenlp} that \emph{retrieve} raw-text from the dialogue history and \emph{construct} the KV cache for each query. RAG-2K retrieves relevant 2K-token chunks, while RAG-Episodic clusters the history into episodes (\Cref{sec:episodic}) and retrieves segments from the most relevant episode under the cache budget $M$.

(2) KV Cache Compression: We compare different compression strategies \emph{within} the block prefill eviction framework under the constrained memory budgets. We include recent query-agnostic compression methods, KVzip~\citep{kim2025kvzipqueryagnostickvcache} and OracleKV~\citep{zhu2025oraclekv}, and additionally compare representative KV cache compression approaches StreamingLLM~\citep{xiao2024efficient}, SnapKV~\citep{li2024snapkv}, KeyDiff~\citep{park2025keydiffkeysimilaritybasedkv}, and InfiniPot~\citep{kim-etal-2024-infinipot}. Detailed dataset information, baseline setups, and \methodname\ configurations are provided in Appendix~\ref{appn:exp_settings}.

\input{figs/fig5/fig5.5}

\input{figs/fig5.6/fig5.6}
\input{figs/fig5.7/fig5.7}

\subsection{Main Results} \label{subsec:main_results}
\Cref{fig:main_exp} presents the main LongConvQA results on three benchmarks (LongMemEval, Realtalk, and LoCoMo) across different KV cache budgets with Qwen3-4B and 8B models. \methodname\ consistently achieves the best performance across all cache budgets and benchmarks, outperforming both retrieval-based baselines (RAG-2K and RAG-Episodic) and recent KV cache compression baselines (KVzip and OracleKV) by wide margins, especially with low cache budgets (2-4K). Moreover, comparing \methodname\ with and without sensitivity-aware budget allocation (\methodname\ w.o. alloc.) shows that our allocation strategy provides a consistent accuracy gain across models and cache budgets. To further validate generality, \Cref{fig:main_realtalk} reports results on Realtalk with four different LLMs, where \methodname\ substantially outperforms a broader set of KV cache compression methods (StreamingLLM, SnapKV, KeyDiff, and InfiniPot), yielding up to 30 absolute score improvement under constrained budgets, while converging to full KV performance as the budget increases. 

\Cref{fig:main_longbench} probes whether the core mechanism of \methodname---clustering semantically coherent segments and retrieving the most relevant episode at inference---transfers beyond dialogue to unstructured document contexts, using four long-form document QA tasks from LongBench~\citep{bai-etal-2024-longbench} (NarrativeQA, HotpotQA, MuSiQue, and QMSum; average input length 15K--36K tokens). \methodname\ consistently outperforms all baselines across both Qwen3-4B and 8B, with the largest gains observed under constrained budgets (2--4K), confirming that episodic KV cache management generalizes effectively to document-level contexts beyond multi-turn conversational structure.

\input{Table/episode}

\subsection{\methodname\ Analysis} \label{sec:ablation}

\paragraph{Multi-Episodic Reasoning.}
We evaluate \methodname's ability to handle queries whose evidence spans multiple conversational episodes from two complementary perspectives. \Cref{fig:main_source_num} leverages the ground-truth indices of the conversation segments provided by RealTalk~\citep{lee2025realtalk}---which indicate the specific multi-turn segments required to answer each query---to regroup queries by the number of source segments referenced (\textit{Src 1/2/3+}), where larger groups demand evidence distributed across more distinct episodes. \Cref{tab:cross_episode} takes an orthogonal view by categorizing each query as \textit{Single}, \textit{Cross}, or \textit{Temporal} based on the sharpness of its cosine similarity distribution over episode centroids obtained from our clustering stage: queries whose similarity is dominated by a single centroid are labeled Single, while those with comparable similarity scores across multiple centroids are labeled Cross; Temporal queries, originally defined in Realtalk, require temporal cues that frequently span episode boundaries.

Both analyses consistently show that \methodname\ maintains strong performance as evidence becomes more distributed across episodes, whereas baselines degrade sharply. In \Cref{fig:main_source_num}, \methodname\ consistently stays closer to Full KV accuracy across all source segment groups (Src~1 to Src~3+), demonstrating robustness to multi-episode evidence that holds across both Qwen3-4B and 8B. \Cref{tab:cross_episode} further confirms this at the query-type level: on RealTalk at 8K, \methodname\ achieves $50.8$ / $55.7$ on Cross and Temporal queries versus RAG's $31.9$ / $22.4$. This robustness stems from a fundamental difference in how context is retained: retrieval-based methods concatenate a small set of retrieved segments at query time, inherently missing dependencies distributed across the full dialogue, while \methodname\ computes each episodic KV cache via block-wise prefill over the \emph{entire} conversation history before compression—preserving globally contextualized representations within every episodic cache. Even when evidence spans multiple episodes, \methodname\ can surface cross-episode dependencies through these context-aware KV states, rather than relying on retrieval to explicitly identify and assemble relevant segments.

\input{Table/ablation}
\input{figs/fig5.8/fig5.8}

\input{Table/efficiency}
\paragraph{Ablation Study.}
\Cref{tab:ablation} ablates design choices of \methodname\ on Realtalk.
We first compare conversation segmentation units : utterance-level splitting consistently outperforms word- and token-level alternatives, confirming that breaking natural speaker-turn boundaries degrades final accuracy.
For embedding model, we compare the LLM's embedding layer, a sentence encoder (MiniLM), and Qwen3-Embedding~\citep{qwen3embedding}. Qwen3-Embedding achieves the best accuracy, while scaling the encoder to 4B yields only marginal gains, indicating that a lightweight encoder (0.6B) is sufficient with negligible overhead relative to the base LLM. We next sweep the number of episodes $E$ and the sharpness parameter $\alpha$ for layer-wise budget allocation. Increasing $E$ improves performance, but even a small number ($E=4$) delivers strong accuracy (49.8 vs.\ 36.0 for KVzip), suggesting that cross-episodic memory is effectively captured in the contextualized KV cache. For sharpness, larger values lead to increasingly skewed layer-wise budgets; moderate sharpness (2--4) performs best, while overly sharp allocation (8) hurts accuracy.

\paragraph{Robustness of Budget Allocation.}
\Cref{fig:main_calib} examines \methodname's robustness across two axes: stochasticity (K-Means initialization and generation sampling) and calibration sensitivity (domain choice and sample count for layer-wise budget profiling).
Across both LoCoMo and Realtalk, all conditions yield consistent performance with standard deviations below $0.5$, confirming that neither clustering randomness nor generation sampling introduces meaningful instability. For calibration, we swept three domains—dialogue history (SAMsum~\cite{gliwa-etal-2019-samsum}), code repository (RepoBench-P~\cite{liu2023repobench}), and long-form narrative (BookSum~\cite{kryściński2022booksumcollectiondatasetslongform})—with $n{\in}\{1, 100\}$ samples each. Profiling with a single sample matches the performance of using 100 samples across all domains, demonstrating that layer-wise sensitivity is largely model-dependent rather than data-dependent: a single offline calibration suffices regardless of domain.

\subsection{Efficiency Analysis}
\label{sec:efficiency}

\Cref{tab:efficiency} reports an efficiency analysis aligned with the pipeline in \Cref{fig:main_method}, measuring the latency of stages 1--3 of \methodname\ and comparing it against two Full KV baselines in a long-term conversational setting under a single-batch inference regime to reflect device-oriented usage. All experiments use a 90K-token dialogue history, corresponding to approximately 30 days of conversational sessions from LongMemEval~\cite{wu2025longmemeval}. Conditioned on this history, we evaluate peak GPU memory usage and total latency over a subsequent 300 user--LLM interactions, and report the latency of each sub-stage along with its execution count.

\methodname\ reduces peak GPU memory by nearly $4\times$ (36.3 vs.\ 9.6), enabling long-term conversational inference under strict memory budgets in mobile environments~\cite{liu2024mobilellmoptimizingsubbillionparameter}. 
For latency, the total cost is largely determined by how often each stage is executed.  Without caching, Full KV re-prefills the entire history at every turn, resulting in a prohibitively high total latency (9,339~s). While prefix caching~\cite{zheng2024sglang} removes repeated prefill, decoding still operates over the full KV cache at each turn, leading to a high latency dominated by decoding (1,062.8~s).

In contrast, although \methodname\ incurs one-time costs for conversation clustering and episodic KV cache construction (stage 1-2 of \Cref{tab:efficiency}), \methodname\ enables decoding to operate  on a compressed KV cache at every turn, achieving up to $2.4\times$ faster decoding, reducing the total latency over long interactions. \methodname\ achieves a total latency of 545.4~s, yielding nearly an $18\times$ speedup over Full KV and an approximately $2\times$ speedup over prefix caching, while maintaining comparable QA accuracy (46.2 vs.\ 45.6).

%% file: figs/fig5/fig5.5.tex
\begin{figure}[t]
    \centering
    \includegraphics[width=0.97\columnwidth]{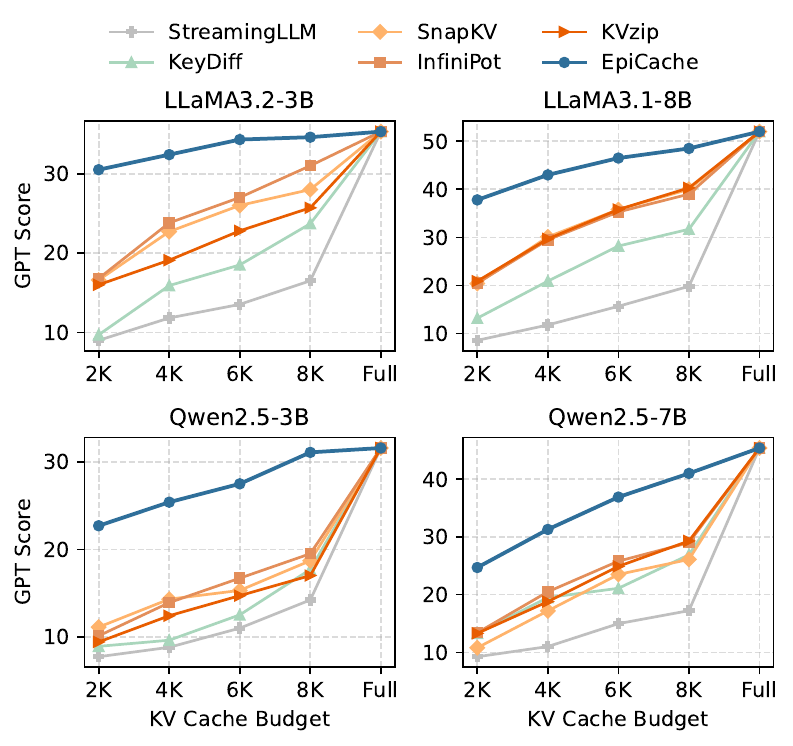}
    \caption{\textbf{Evaluation with Various Models} results with Realtalk. Evaluated with fixed KV cache budgets $M\in\{2\text{K},4\text{K},6\text{K},8\text{K}\}$ across four LLMs. The number of episodes (clusters) fixed to $E{=}4$ in all experiments. See \Cref{fig:appn_main_convqa} for other two benchmarks.}
    \label{fig:main_realtalk}
    \vspace{-0.15in}
\end{figure}

%% file: figs/fig5.6/fig5.6.tex
\begin{figure}[t]
    \centering
    \includegraphics[width=0.95\columnwidth]{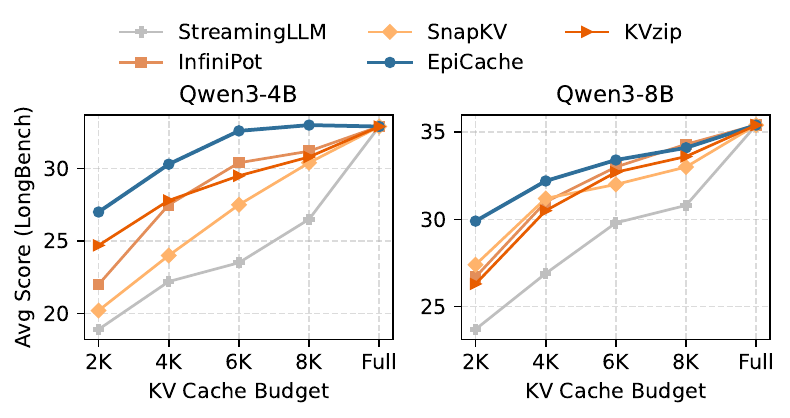}
    \caption{\textbf{Long Form Document QA Results} on NarrativeQA, HotpotQA, MuSiQue, and QMSum (averaged). Evaluated with fixed KV cache budgets $M\in\{2\text{K},4\text{K},6\text{K},8\text{K}\}$ on Qwen3-4B and Qwen3-8B. The number of episodes fixed to $E{=}4$.}
    \label{fig:main_longbench}
    \vspace{-.2in}
\end{figure}

%% file: figs/fig5.7/fig5.7.tex
\begin{figure}[t]
    \centering
    \includegraphics[width=0.97\columnwidth]{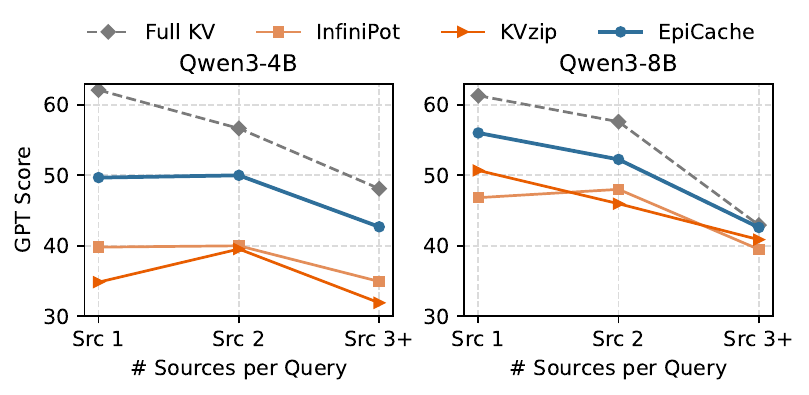}
\caption{\textbf{Multi-Episode Analysis} on RealTalk with Qwen3-4B and 8B at budget $M{=}8\text{K}$. Queries are grouped by the number of sources required to answer (\textit{Src 1/2/3+}), where larger groups demand evidence distributed across more episodes.}
    \label{fig:main_source_num}
\end{figure}

%% file: Table/episode.tex
\begin{table}[t]
\centering
\resizebox{\columnwidth}{!}{
\begin{tabular}{l|c|ccc|ccc}
\toprule
{Qwen3-4B}
& & \multicolumn{3}{c|}{\textbf{RealTalk}} 
& \multicolumn{3}{c}{\textbf{LoCoMo}} \\
\cmidrule(lr){1-2}\cmidrule(lr){2-3}\cmidrule(lr){3-5}\cmidrule(lr){6-8}
Method & $M$ & {Single} & {Cross} & {Temp.}
& {Single} & {Cross} & {Temp.} \\
\midrule
Full KV & - & 60.3 & 55.4 & 61.7 & 34.1 & 32.8 & 32.6 \\
\midrule
RAG-2K   & 4K & 31.8 & 27.9 & 17.6 & 24.2 & 20.0 & 5.5 \\
RAG-2K  & 8K & 38.4 & 31.9 & 22.4 & 31.6 & 27.8 & 7.3 \\ \midrule
KVzip & 4K & 26.7 & 21.8 & 22.8 & 13.7 & 15.7 & 13.7 \\
KVzip & 8K & 37.5 & 35.9 & 35.0 & 24.5 & 24.3 & 23.0 \\
\midrule
\rowcolor{gray!8} \methodname & 4K & 45.3 & 40.3 & 41.5 & 29.8 & 25.8 & 22.3 \\
\rowcolor{gray!16} \methodname & 8K & 54.6 & 50.8 & 55.7 & 33.3 & 32.7 & 29.2 \\
\bottomrule
\end{tabular}
}
\vspace{.05in}
\caption{\textbf{Cross Episode Analysis} on RealTalk and LoCoMo with Qwen3-4B under KV cache budgets $M\in\{4\text{K},8\text{K}\}$.}
\vspace{-.25in}
\label{tab:cross_episode}
\end{table}

%% file: Table/ablation.tex
\begin{table}[t]
\centering
\resizebox{\columnwidth}{!}{
\begin{tabular}{lccc|c}
\toprule
Qwen3-4B ($M$=8K) & \textbf{Multi-Hop} & \textbf{Temporal} & \textbf{Common} & \textbf{Avg} \\
\midrule
Full KV        & 53.6 & 61.7 & 52.2 & 56.9 \\ \midrule
RAG-Episodic & 42.3 & 22.4 & 41.0 & 33.4 \\
KVzip       & 34.4 & 35.0 & 43.3 & 36.0 \\ \midrule
\multicolumn{5}{l}{\textit{Conversation segmentation unit} ($E=4$, w.o budget allocation)} \\
Words               & 44.4 & 50.3 & 47.6 & 47.5 \\
Tokens             & 47.9 & 47.9 & 49.3 & 47.9 \\
\rowcolor{gray!8} Utterances  & 48.5 & 51.5 & 48.1 & 49.8 \\
\midrule
\multicolumn{5}{l}{\textit{Embedding model $f_{embed}$} ($E=4$, w.o budget allocation)} \\ 
LLM-embedding  & 40.8 & 43.1 & 49.1 & 43.0 \\
MiniLM-L6-v2   & 45.5 & 47.9 & 52.2 & 47.6 \\
\rowcolor{gray!8} Qwen3-Emb-0.6B & 48.5 & 51.5 & 48.1 & 49.8 \\
Qwen3-Emb-4B   & 48.6 & 53.1 & 48.8 & 50.6 \\
\midrule
\multicolumn{5}{l}{\textit{The number of episodes $E$} (w.o budget allocation)} \\
E = 2          & 46.5 & 50.0 & 46.9 & 47.9  \\
\rowcolor{gray!8} E = 4 & 48.5 & 51.5 & 48.1 & 49.8 \\
E = 6          & 48.0 & 56.1 & 50.8 & 51.1 \\
E = 8          & 48.5 & 53.4 & 53.1 & 51.3 \\
\midrule
\multicolumn{5}{l}{\textit{Layer-wise budget allocation (sharpness hyper-parameter $\alpha$)}} \\
$\alpha = 1$   & 49.0 & 55.3 & 54.4 & 53.0 \\
\rowcolor{gray!8} $\alpha = 2$ & 51.7 & 55.7 & 54.7 & \textbf{53.9} \\
$\alpha = 4$   & 50.2 & 57.4 & 54.1 & 53.9 \\
$\alpha = 8$   & 48.3 & 52.1 & 47.3 & 49.8 \\
\bottomrule
\end{tabular}
}
\vspace{.05in}
\caption{\textbf{Ablation Study} on the {RealTalk} benchmark with Qwen3-4B 8K cache budget. LLM-embedding uses the target LLM's embedding layer as the encoder, and Qwen3-Emb. denotes the Qwen3 embedding model. Highlighted rows indicate the final selected configuration of \methodname.}
\vspace{-.25in}
\label{tab:ablation}
\end{table}

%% file: figs/fig5.8/fig5.8.tex
\begin{figure}[t]
    \centering
    \includegraphics[width=1\columnwidth]{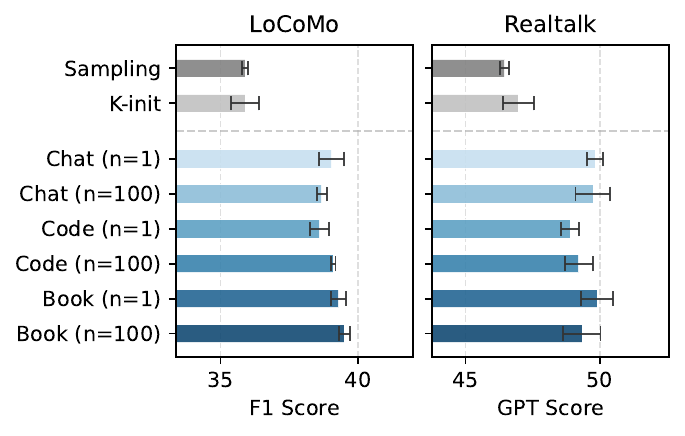}
\caption{\textbf{Robustness of Layer-wise Budget Allocation} on LoCoMo and Realtalk with Qwen3-8B at budget $M{=}6\text{K}$. Error bars denote $\pm$std over 5 seeds. \textit{Top}: variance due to K-Means initialization (K-init) and generation sampling, evaluated without budget allocation. \textit{Bottom}: variance across layer sensitivity profiling calibration domain and number of calibration samples ($n{=}1$ or $100$).}
    \label{fig:main_calib}
    \vspace{-0.2in}
\end{figure}

%% file: Table/efficiency.tex
\begin{table*}[t]
\centering
\resizebox{\textwidth}{!}{
\begin{tabular}{l|l|cc|c|ccc|c|c|c}
\toprule
{LLaMA3.2-3B} & 300 Turns & \multicolumn{2}{c|}{\textbf{Stage 1}} & \textbf{Stage 2}
 & \multicolumn{3}{c|}{\textbf{Stage 3}} 
 & Total & QA & Peak \\
{Method} 
& w. 90K History
& Embed
& Cluster
& Prefill 
& Match
& Retr. \& wb
& Decode 
& (Per-Turn)
& Acc.
& Mem. (GB)
\\
\midrule

\multirow{2}{*}{{Full KV}}
& Latency (sec)
& 0.0
& 0.0
& 27.8
& 0.0
& 0.0
& 3.5
& \multirow{1}{*}{9339.0}
& \multirow{2}{*}{46.2}
& \multirow{2}{*}{36.3}
\\
& \# of Execution
& 0
& 0
& $\times$300
& 0
& 0
& $\times$300
& (31.1)
& 
& 
\\

\midrule

\multirow{1}{*}{{Full KV}}
& Latency (sec)
& 0.0
& 0.0
& 27.8
& 0.0
& 0.0
& 3.5
& \multirow{1}{*}{1062.8}
& \multirow{2}{*}{46.2}
& \multirow{2}{*}{36.3}
\\
w. Prefix Caching
& \# of Execution
& 0
& 0
& $\times$1
& 0
& 0
& $\times$300
& (3.5)
& 
& 
\\

\midrule



\multirow{2}{*}{\textbf{\methodname}}
& Latency (sec)
& 25.6
& 0.76
& 18.0
& 0.04
& 0.30
& 1.4
& \multirow{1}{*}{\textbf{545.4}}
& \multirow{2}{*}{45.6}
& \multirow{2}{*}{\textbf{9.6}}
\\
& \# of Execution
& $\times$1
& $\times$1
& $\times$4
& $\times$1
& $\times$90
& $\times$300
& (1.8)
& 
& 
\\

\bottomrule
\end{tabular}
}
\vspace{0.05in}
\caption{
\textbf{Efficiency Analysis} with GPU memory usage (GB) and Latency (sec), under 30 days of 90K-token conversation history followed by subsequent 300 multi-turn conversation with LLaMA3.2-3B. $M=8$K cache budget used for \methodname. Retr. \& wb (retrieval and write-back) stage is executed 90 times over 300 turns (30\% topic-change rate), and its cost is amortized accordingly. }
\vspace{-.1in}
\label{tab:efficiency}
\end{table*}

%% file: Section/Related.tex
\section{Related Work}

Prior KV cache eviction methods rely on attention scores to estimate token importance. H2O~\cite{zhang2023ho} aggregates entire attention scores for eviction, while SnapKV~\citep{li2024snapkv} uses query-dependent attention scores. InfiniPot~\citep{kim-etal-2024-infinipot} and OracleKV~\citep{zhu2025oraclekv} use designed patched prompts that approximate future queries to derive importance scores. Alongside attention-based eviction, TRIM-KV~\citep{bui2025cachelaststokenretention} learns token importance via a lightweight retention gate fine-tuned through distillation, and Cartridge~\citep{eyuboglu2025cartridgeslightweightgeneralpurposelong} trains a corpus-specific KV cache offline using a self-study objective, both requiring additional training. FlowKV~\citep{liu2025flowkv} introduces a multi-turn isolation mechanism that avoids recompressing past-turn KV caches, but does not compress conversation history.

Another line of work improves decoding efficiency by selectively retrieving the most relevant parts of the KV cache for each query tokens, thereby reducing the cost of attention computation. Quest~\citep{tang2024quest} and ArkVale~\citep{chen2024arkvale} retrieves KV entries at the granularity of pages, while SqueezedAttention~\citep{hooper-etal-2025-squeezed} and ClusterKV~\citep{liu2024clusterkvmanipulatingllmkv} clusters Key states and loads the cluster relevant to the query. IceCache~\citep{mao2026icecache} further extends this line by combining semantic token clustering with PagedAttention to hierarchically organize KV cache. These methods share two key limitations. First, they operate in the \textit{post-prefill} regime, assuming unconstrained memory usage during prefill. Second, their retrieval units (pages and cluster) do not align with the episodic structure of conversations, limiting their applicability to LongConvQA under strict memory budgets.

%% file: Section/Conclusion.tex
\section{Conclusion}
\methodname\ is the first framework that combines block-wise prefill with episodic clustering and sensitivity-aware budget allocation to preserve topic-relevant context under a fixed memory budget. Across multiple LongConvQA benchmarks, \methodname\ substantially outperforms existing compression methods, demonstrating that long-term interaction is feasible under strict resource constraints and marking a practical step toward on-device conversational AI.

\paragraph{Future Work.} Future work includes improving episode construction through more effective conversational embeddings~\citep{yang-etal-2025-learning} and adaptive episode count selection~\citep{clustering_dialogue}. Evaluating \methodname\ on more challenging benchmarks that require tracking implicit user preferences and handling information updates
over time~\citep{jiang2025personamemv2personalizedintelligencelearning, uddin2026recallforgettingbenchmarkinglongterm} presents a natural next step toward long-term conversational agents.

\section*{Impact Statement}
This paper presents work whose goal is to advance the field of Machine Learning. There are many potential societal consequences of our work, none which we feel must be specifically highlighted here.

%% file: Section/Appendix.tex
\newpage
\appendix
\onecolumn

\renewcommand{\thefigure}{A\arabic{figure}}
\renewcommand{\thetable}{A\arabic{table}}
\setcounter{figure}{0}

\section{Experimental Details}\label{appn:exp_settings}
\subsection{Dataset}\label{appn:dataset}
We evaluate \methodname\ on three LongConvQA benchmarks: Realtalk~\citep{lee2025realtalk}, LoCoMo~\citep{maharana-etal-2024-evaluating}, and LongMemEval~\citep{wu2025longmemeval}. Three benchmarks follow the LongConvQA formulation in \Cref{sec:longconvqa}, where a long conversational history $\mathcal{H}$ is provided and the model is required to answer a sequence of queries $\mathcal{Q} = {q_1, \dots, q_{N_q}}$ grounded in dialogue history. This formulation evaluates the answer accuracy of LLMs in a \textit{multi-turn conversation}.

\paragraph{Realtalk.}
Realtalk~\citep{lee2025realtalk} is a real-world dataset of 10 long-term conversations, where pairs of participants engaged in daily messaging for 16-21 days. Unlike LLM-simulated corpora such as LoCoMo, Realtalk captures natural dialogue including typos, abbreviations, asynchronous response gaps, and consecutive messages, while also reflecting diverse emotional expressions and shifts in persona consistency. 

For evaluation, the dataset provides annotated memory probing questions across three subtasks-multi-hop, temporal reasoning, and commonsense—requiring models to recall and reason over extended histories. Following the original setup, we adopt GPT-based scoring (\texttt{gpt-4o-mini-2024-07-18}) to assess open-ended generation.

\paragraph{LoCoMo.} LoCoMo~\citep{maharana-etal-2024-evaluating} is a benchmark of long-term conversations, created through a human-machine pipeline where LLM-based agents generate dialogues grounded in distinct personas and temporal event graphs, and human annotators refine them for long-range consistency. The dataset consists of 10 conversations, each spanning up to 35 sessions with around 300 turns. The QA benchmark is divided into five subtasks: (i) Single-hop, (ii) Multi-hop, (iii) Temporal reasoning, (iv) Open-domain knowledge, and (v) Adversarial. Evaluation is conducted with open-ended genetation with F1 score.

We exclude the \textit{adversarial} subtask for the following reason. This task tests whether a model can recognize unanswerable questions by choosing between a plausible but incorrect answer and a ``no such information'' response. Under KV cache compression, however, models frequently over-predict the latter, which leads to spuriously high scores. For example, with LLaMA-3.2-3B the adversarial score is only 12.11 under full KV, yet jumps to 49.78 with 4K KVzip compression—an increase that reflects bias rather than genuine improvement. 

This behavior contrasts with other subtasks such as temporal or multi-hop reasoning, where compressed caches consistently degrade performance. Because open-source models already struggle on adversarial questions~\citep{maharana-etal-2024-evaluating}, reporting these inflated numbers would give a misleading evaluation of answer quality under compressed KV cache. We therefore omit adversarial results from our main evaluation and defer a more thorough study of unanswerability detection under compression to future work.

\paragraph{LongMemEval.} LongMemEval~\citep{wu2025longmemeval} benchmarks long-term memory in user-assistant interactions with five core abilities—information extraction, multi-session reasoning, temporal reasoning, knowledge updates, and abstention—through seven question types (single-session user/assistant/preference, two-hop, multi-session synthesis, knowledge update, temporal reasoning, and abstention). A key property is its \emph{length-configurable} chat histories: the benchmark provides standardized settings with extremely long contexts (e.g., up to 1.5M tokens), designed as controlled stress tests of memory and retrieval mechanisms. We follow the open-ended generation setup and report F1 scores for this dataset.  

To align LongMemEval with the LongConvQA formulation in \Cref{sec:longconvqa}, we utilize the custom session stacking provided by LongMemEval\footnote{\url{https://github.com/xiaowu0162/LongMemEval}} to build coherent long conversations from user-LLM. Using this feature, we construct evaluation sets while preserving the original distribution of all question types. Specifically, we sample QA pairs according to the benchmark's task-type proportions, retrieve the corresponding evidence conversation sessions, and assemble them into chronologically consistent histories. We then evaluate models at context lengths of 20K, 40K, 60K, 80K, and 100K tokens. This design allows us to test KV cache compression under scalable memory budgets.

\input{Table/appn_algorithm}

\subsection{KV Cache Compression Baseline Setup}\label{appn:baseline}

We adapt existing KV cache compression methods to the block prefill setting for a fair comparison with our approach. 
The baselines include both static retention and attention-based eviction strategies, as well as similarity-based selection.

\paragraph{StreamingLLM.}
Following StreamingLLM~\citep{xiao2024efficient}, we retain a fixed number of sink and recent tokens throughout block prefill. 
Specifically, we fix the number of sink tokens to 128 for all models, while the remaining budget $M-128$ is assigned to the most recent tokens. 

\paragraph{SnapKV.}
We adapt SnapKV~\citep{li2024snapkv} to the block prefill setting, where future queries are not accessible in LongConvQA. Following the original design, we use the window tokens—in our case the last 64 tokens of each block—as the patched prompt, and then apply the scoring function in \Cref{eq:patched_prompt}. Tokens with the highest attention relevance to this patched prompt are retained.

\paragraph{KVzip.}
We adapt KVzip~\citep{kim2025kvzipqueryagnostickvcache} to block prefill by treating the entire block of tokens as the patched prompt. At each block boundary, we append a repetition instruction (e.g., ''Repeat the part of the previous context exactly") followed by the full block tokens, and then apply the patched-prompt scoring method. 

\paragraph{InfiniPot.}
We adopt the InfiniPot~\citep{kim-etal-2024-infinipot} by employing a general-purpose patched prompt designed to highlight globally important content. 
Specifically, we append the instruction 
``Summarize the previous context highlighting the most important parts." 
at the end of each block and compute scores according to \Cref{eq:patched_prompt}. 
This encourages selection of semantically informative tokens across the block.

\paragraph{KeyDiff.}
KeyDiff~\citep{park2025keydiffkeysimilaritybasedkv} is KV cache eviction method for block prefill. For each block, it constructs an anchor key by averaging the key states of all tokens, and computes the dissimilarity score of each token as the negative cosine similarity between its key state and the anchor. These scores are then used to guide eviction. Following the original implementation, we evaluate $M_{\text{block}} \in {128, 512, 1024, 2048}$, which includes the default setting of 128, and report results using the configuration that achieved the best performance.

To ensure fair comparison, all attention-based methods (SnapKV, InfiniPot, KVzip, \methodname) use the same scoring formulation from \Cref{eq:patched_prompt}, and all eviction methods are combined with head-wise non-uniform token selection as suggested by \citet{feng2024ada}. 

\subsection{\methodname\ Setup}\label{appn:ours}
\paragraph{Overall Process.} We provide the complete procedure of \methodname\ in \Cref{alg:epicache_all}. The framework consists of two phases. In \textbf{Phase A}, the conversation history is clustered into topical episodes and a compressed KV cache is prepared for each episode. In \textbf{Phase B}, online queries are answered by retrieving the most relevant episodic cache. 

In \textit{\textcolor{teal}{Phase A1}}, we segment the conversation history, embed each segment, and cluster them into $E$ episodes. This step can be performed offline, and the cost of segment encoding and K-means clustering is negligible (under a minute). Each episode is represented by its centroid and a medoid segment that serves as a patched prompt.
In \textit{\textcolor{teal}{Phase A2}}, we measure per-layer sensitivity by comparing Key states under full and block-wise prefill masks, and allocate layer-wise budgets proportionally using the sharpness hyper-parameter $\alpha$. In \textit{\textcolor{teal}{Phase A3}}, we construct episodic KV caches by performing block-wise prefill with the patched prompt appended, and then compress the resulting caches according to the allocated budgets from Phase A2. Although prefill must be repeated for every episode, the peak memory remains flat during this process (see \Cref{fig:prelim_c}), making it practical for constrained-memory environments.

In Phase B, when a new query arrives, we embed the query and compute its similarity to the episode centroids. The query is then routed to the most relevant episodic cache, which is loaded for decoding. If the same cache is selected as in the previous turn, no additional retrieval is required since the cache remains resident, further reducing overhead. 

\paragraph{Detailed Settings} For segment construction, we set the embedding window size $w_{\text{embed}}$ to 4, selected from ${2, 4, 8}$. To cluster segments into episodes, we apply the standard K-means algorithm with k-means++ initialization~\citep{10.5555/1283383.1283494}. This offline segmentation and clustering stage completes within one minute, incurring negligible overhead.

For sensitivity-aware budget allocation, we estimate per-layer weights using a single randomly sampled long document from the BookSum~\citep{kryściński2022booksumcollectiondatasetslongform} dataset, chosen to avoid bias from any specific conversational dataset. By performing two forward passes—one with the full causal mask and one with the block-wise prefill mask—we measure layer-wise deviations and compute allocation weights. Because only one sample is used, the overhead of this calibration step is negligible.

In block prefill, the cache always maintains size $M$: as conversation segments are added in blocks of $M_{\text{block}}$, the cache can temporarily grow up to $M + M_{\text{block}}$ entries, after which eviction is applied to reduce it back to $M$. A larger $M_{\text{block}}$ enables the model to cover the entire conversation more quickly but increases the temporary peak memory footprint, while a smaller $M_{\text{block}}$ lowers peak memory at the cost of slower coverage. We set $M_{\text{block}} \in {128, 512, 1024, 2048}$ to balance this trade-off.

\section{Further Analysis}\label{appn:analysis}
\input{Table/appn_clustering}
\subsection{Conversation Clustering Analysis}\label{appn:analysis_clustering}

In this section, we provide qualitative examples of conversation clustering to illustrate how episodic structures emerge in practice. Conversation histories are divided into segments of $w_{\text{embed}}=4$ utterances, which are then embedded using Qwen3-0.6B~\citep{qwen3embedding}. Segment embeddings are clustered with K-Means, and the resulting clusters are visualized in two dimensions via t-SNE, as shown in \Cref{fig:tsne_medoids}~(a).

For each cluster, we further present representative medoid segments in \Cref{fig:tsne_medoids}~(b). These examples demonstrate that the clustering procedure consistently groups segments into coherent topical episodes, such as games, movies, literature, or weather. The medoid samples highlight the interpretability of each episode and indicate how such episode-level partitioning can serve as the basis for episodic KV cache compression.

\input{figs/appn_layer/appn_layer}

\subsection{block-wise prefill Sensitivity Analysis}\label{appn:analysis_sensitivity}
We analyze layer-wise deviations under block prefill by comparing multiple internal states—Key, Value, and layer outputs—across Transformer layers computed with the full causal mask ($\mathcal{M}$) and the block mask ($\mathcal{M}'$). To this end, we forward LoCoMo~\citep{maharana-etal-2024-evaluating} conversation history samples under both masks and plot the resulting internal states differences across layers, as defined in \Cref{eq:mask,eq:mask_diff}. Specifically, each plot reports cosine similarity of Key and Value states, and L2 distance of layer outputs, respectively.  

We find that Value states (\Cref{fig:layer_value}) exhibit consistently low similarity across layers, offering little discriminative trend. Layer outputs, measured by L2 distance (\Cref{fig:layer_layer}), show a monotonic error accumulation pattern rather than meaningful variation. In contrast, Key states (\Cref{fig:layer_key}) provide clear differentiation across layers. This observation motivates our use of Key state deviation as the sensitivity measure for budget allocation, as discussed in \Cref{sec:adaptive}. Further analysis of why these trends differ across Key, Value, and layer-wise output representations is left for future work.

\section{Additional Experimental Results}\label{appn:exp_results}

\input{figs/fig6/fig6}

\subsection{Comprehensive LongConvQA Results}\label{appn:longconvqa}
\input{figs/fig5/fig5_appn}
\Cref{fig:appn_main_convqa} shows the comprehensive LongConvQA evaluation results, comparing \methodname\ against baselines~\citep{xiao2024efficient,li2024snapkv,kim-etal-2024-infinipot,park2025keydiffkeysimilaritybasedkv,kim2025kvzipqueryagnostickvcache} across multiple cache budgets and four model variants, LLaMA3.2-3B, LLaMA3.1-8B, Qwen2.5-3B, and Qwen2.5-7B. The corresponding numerical results underlying this figure are reported in \Cref{tab:appn_convqa_llama} for Realtalk and LoCoMo with LLaMA3.2-3B and LLaMA3.1-8B, in \Cref{tab:appn_convqa_qwen} for Qwen2.5-3B and 7B, and in \Cref{tab:lme_side_by_side} for LongMemEval across all four models.

\subsection{Memory Scalability Evaluation.}\label{appn:memory_scale}
\Cref{fig:main_scale} evaluates LongConvQA under extended conversation lengths, scaling up to 100K tokens.\footnote{LongMemEval~\citep{wu2025longmemeval} supports stacking conversation sessions with associated QA pairs, allowing conversation histories to be constructed at custom lengths.} 
Open-source LLMs exhibit declining QA performance as context length grows to 100K, as reported in \cite{wu2025longmemeval}, and the performance gap between full KV and baseline methods (KVzip, InfiniPot) becomes increasingly pronounced. 
\methodname\ delivers higher accuracy than baselines at the same memory budget across all context lengths, and as the KV cache budget increases, its accuracy steadily approaches full KV, demonstrating the memory scalability of our approach.

\section{Future Work}\label{appn:future}

Several promising directions remain for future work. On the episode construction side, the quality of episodic boundaries depends heavily on the underlying segment representations: more effective conversational embeddings that capture turn-level context and inter-utterance dependencies could yield more coherent episode construction~\citep{yang-etal-2025-learning}, representing an improvement to our clustering and scoring pipeline. Complementarily, extending \methodname\ to adaptively determine the optimal number of
episodes~\citep{clustering_dialogue} could further improve scalability across dialogue lengths and domains.

Beyond improvements to episodic construction, long-term conversational QA surfaces richer memory management challenges: tracking implicit user preferences, handling information updates and deletions over time, and maintaining consistent persona across extended interactions~\citep{jiang2025personamemv2personalizedintelligencelearning,uddin2026recallforgettingbenchmarkinglongterm}. Finally, combining text-level memory organization~\citep{tan-etal-2025-prospect,pan2025secom, xu2026amem} with efficient KV cache compression presents a promising direction toward accurate and efficient conversational agents deployable on resource-constrained devices.

\input{Table/appn_longconvqa_llama}
\input{Table/appn_longconvqa_qwen}

\input{Table/appn_longmemeval_llama_qwen}

%% file: Table/appn_algorithm.tex
\begin{algorithm*}[t]
\small
\caption{\small \methodname\ with Layer-wise Budget Allocation Pseudo Code}
\label{alg:epicache_all}
\begin{algorithmic}[1]
\REQUIRE $\mathcal{H}$ (history, $N_u$ turns), $f_{\text{embed}}$, $w_{\text{embed}}$, $E$, $M$; 
$f_{\mathrm{LM}}$ ($L$ layers, $H$ heads); masks $\{\mathcal{M},\mathcal{M}'\}$; 
sharpness $\alpha$; calibration batch $\mathcal{B}$ with $|\mathcal{B}|=1$; 
patched prompt $\mathcal{P}_e$ (built from medoid segments, see stage A1)
\ENSURE Episodic caches $\mathbb{B}=\{C_{\mathrm{KV}}^{(1)},\dots,C_{\mathrm{KV}}^{(E)}\}$, centroids $\{\mathbf{c}_e\}_{e=1}^{E}$, and layer budgets $\{M^{\mathrm{alloc}}_{\ell}\}_{\ell=1}^{L}$

\item[] \textbf{Phase A: Clustering and Prefill}

\item[] {\textcolor{teal}{\textit{A1. Conversation Segment \& Clustering (Offline)}}}
\STATE Partition $\mathcal{H}$ into $K=\lceil N_u/w_{\text{embed}}\rceil$ segments $\{S_k\}_{k=1}^{K}$ and encode $\mathbf{e}_k=f_{\text{embed}}(S_k)$.
\STATE Run K-Means clustering on $\{\mathbf{e}_k\}$ to obtain $\{\mathcal{E}_e\}_{e=1}^{E}$.
\FOR{$e=1$ \textbf{to} $E$}
  \STATE $\mathbf{c}_e \gets \frac{1}{|\mathcal{E}_e|}\sum_{S_k\in\mathcal{E}_e}\mathbf{e}_k$;\quad
  $S^{(e)}_{\mathrm{medoid}} \gets \arg\max_{S_k\in\mathcal{E}_e}\cos(\mathbf{e}_k,\mathbf{c}_e)$
  \STATE Build patched prompt $\mathcal{P}_e$ by concatenating utterances of $S^{(e)}_{\mathrm{medoid}}$.
\ENDFOR

\item[] {\textcolor{teal}{\textit{A2. Measure layer sensitivity \& allocate KV budgets.}}}
\FOR{each $x \in \mathcal{B}$}
  \STATE $K^{\ell}_{\mathrm{full}}(x)\gets f_{\mathrm{LM}}(x,\mathcal{M})W_K^{\ell}$;\quad
        $K^{\ell}_{\mathrm{block}}(x)\gets f_{\mathrm{LM}}(x,\mathcal{M}')W_K^{\ell}$ \textbf{for} $\ell=1{:}L$
  \STATE $\sigma_\ell(x)\gets\frac{1}{HN}\sum_{h=1}^{H}\sum_{i=1}^{N}\cos\!\big(k^{(\ell,h)}_{\mathrm{full},i}(x),k^{(\ell,h)}_{\mathrm{block},i}(x)\big)$
\ENDFOR
\STATE $\sigma_\ell \gets \frac{1}{|\mathcal{B}|}\sum_{x\in\mathcal{B}}\sigma_\ell(x)$;\;\; $s_\ell \gets 1-\sigma_\ell$
\STATE $w_\ell \gets \dfrac{s_\ell^{\,\alpha}}{\sum_{j=1}^{L}s_j^{\,\alpha}}$;\;\; $M^{\mathrm{alloc}}_{\ell} \gets (L\cdot M)\,w_\ell$

\item[] {\textcolor{teal}{\textit{A3. Build episodic KV caches.}}}
\FOR{$e=1$ \textbf{to} $E$}
  \STATE Block-wise prefill over $\mathcal{H}$, appending $\mathcal{P}_e$ to each block of $M_{\text{block}}$ tokens.
  \STATE Compute scores w.r.t. $\mathcal{P}_e$ with \Cref{eq:patched_prompt} and retain the top $M$ tokens.
  \STATE $C_{\mathrm{KV}}^{(e)} \gets$ compressed cache for episode $e$.
\ENDFOR
\STATE $\mathbb{B} \gets \{C_{\mathrm{KV}}^{(1)},\dots,C_{\mathrm{KV}}^{(E)}\}$.

\item[] \textbf{Phase B: Online decoding}
\STATE For query $q_i$: $\mathbf{q}_i \gets f_{\text{embed}}(q_i)$;\quad
$e^\dagger \gets \arg\max_{e}\cos(\mathbf{q}_i,\mathbf{c}_e)$
\STATE Retrieve $C_{\mathrm{KV}}^{(e^\dagger)}$ and generate with compressed cache:
$f_{\mathrm{LM}}(q_i \mid C_{\mathrm{KV}}^{(e^\dagger)})$.
\end{algorithmic}
\end{algorithm*}

%% file: Table/appn_clustering.tex
\begin{figure*}[t]
\centering
\begin{minipage}[t]{0.40\linewidth}
  \vspace{0pt} 
  \centering
  \includegraphics[width=\linewidth]{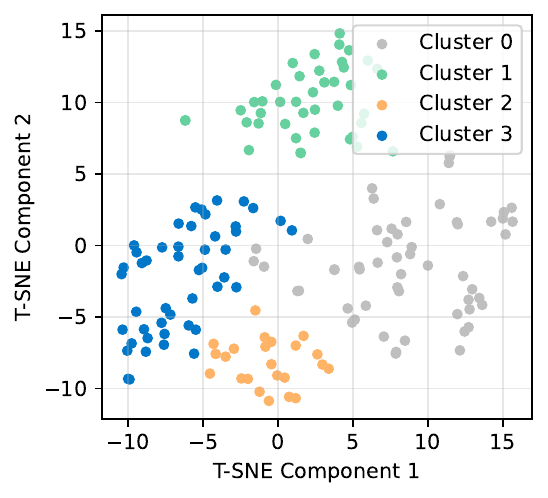}
  \caption*{(a) \small t-SNE visualization} \label{fig:tsne}
\end{minipage}\hfill
\begin{minipage}[t]{0.58\linewidth}
  \vspace{0pt} 
  \raggedright
  \small
  \begin{tabular}{@{}p{0.24\linewidth} p{0.7\linewidth}@{}}
  \toprule
  \textbf{Episode} & \textbf{Medoid segments examples} \\
  \midrule
  \textbf{0. Video game} &
  A: Its a game I used to play a lot … mostly play for fun … \newline
  B: I also have finished the first game … my favorite games of all time are ... \\ \midrule
  \textbf{1. Movie} &
  A: I love the art and I think he is an incredible director …  \newline
  B: I haven’t watched movies in a while … \\ \midrule
  \textbf{2. Literature} &
  A: Have you ever read a book called …  \newline
  B: I’m currently reading ... a multi generational family ... \\ \midrule
  \textbf{3. Weather} &
  A: I love the weather today its gotten warmer \newline
  B: I just got home from work … it’s raining like crazy … parts of the city are flooded \\ 
  \bottomrule
  \end{tabular}
  \caption*{\small(b) Medoid samples by cluster.} \label{tab:medoids}
\end{minipage}

\caption{\textbf{Episodic clustering of conversation segments.} 
(a) t-SNE visualization of conversation clustering. (Silhouette score=0.28) (b) Medoid segments illustrate coherent topics per cluster.}
\label{fig:tsne_medoids}
\vspace{-.2in}
\end{figure*}

%% file: figs/appn_layer/appn_layer.tex
\begin{figure*}[t]
    \centering
    \begin{subfigure}[t]{0.32\textwidth}
    \captionsetup{justification=centering}
        \includegraphics[width=\textwidth]{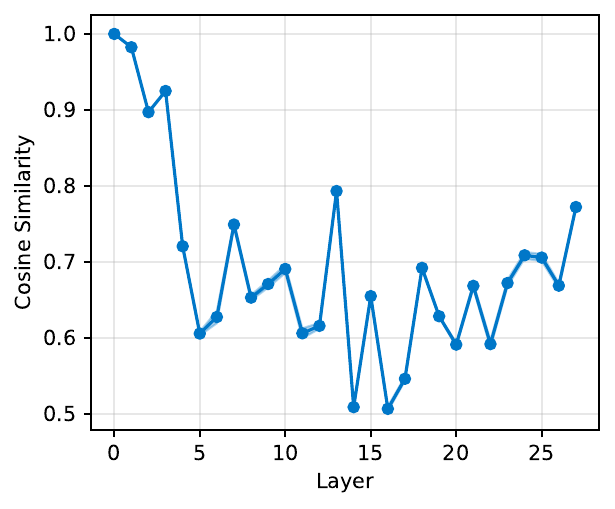}
        \subcaption{Key states similarity}
        \label{fig:layer_key}
    \end{subfigure}
    \centering
    \begin{subfigure}[t]{0.32\textwidth}
    \captionsetup{justification=centering}
        \includegraphics[width=\textwidth]{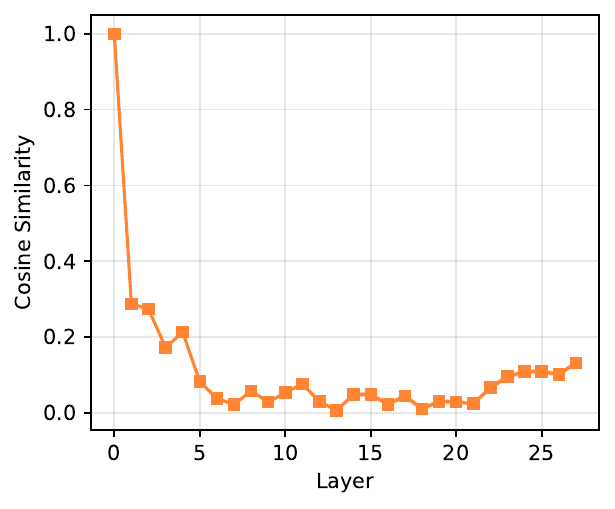}
        \subcaption{Value states similarity}
        \label{fig:layer_value}
    \end{subfigure}
    \centering
    \begin{subfigure}[t]{0.32\textwidth}
    \captionsetup{justification=centering}
        \includegraphics[width=\textwidth]{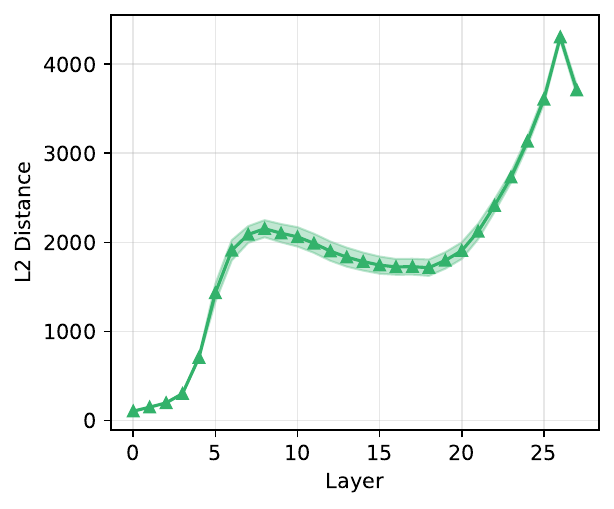}
        \subcaption{Layer outputs L2 distance}
        \label{fig:layer_layer}
    \end{subfigure}
    \caption{\textbf{Layer-wise Sensitivity Analysis.} layer-wise deviation results under the full and block masks using Qwen2.5-7B on LoCoMo conversation history. Key and Value states are measured by cosine similarity, while hidden states at layer outputs are measured by L2 distance; shaded regions indicate variance across input samples.}
    \vspace{-.2in}
    \label{fig:layer_diff}
\end{figure*}

%% file: figs/fig6/fig6.tex
\begin{figure*}[t]
    \centering
    \includegraphics[width=0.98\textwidth]{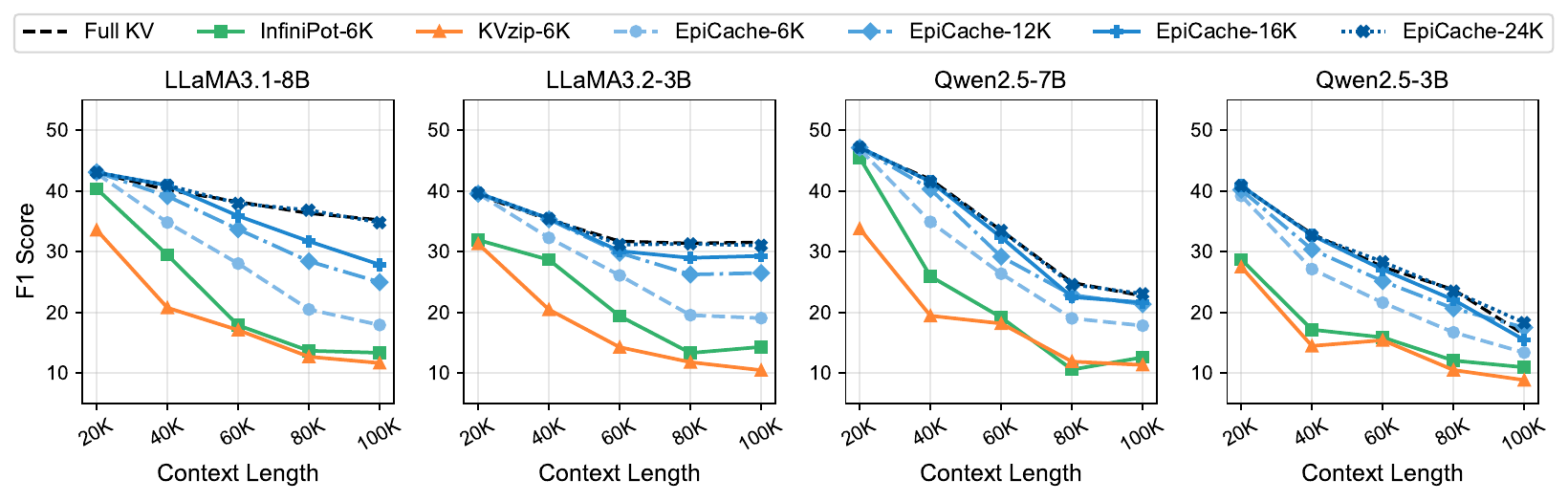}
    \caption{\textbf{Memory Scalability up to 100K Context.} Conversation histories between user and LLM-based assistant scaled to 100K tokens across four LLMs with LongMemEval. Comparison of InfiniPot and KVzip ($M{=}6$K) with \methodname\ (4 episodes, $M{=}6$K–24K).}
    \label{fig:main_scale}
    \vspace{-0.1in}
\end{figure*}

%% file: figs/fig5/fig5_appn.tex
\begin{figure}[t]
    \centering
    \includegraphics[width=0.95\columnwidth]{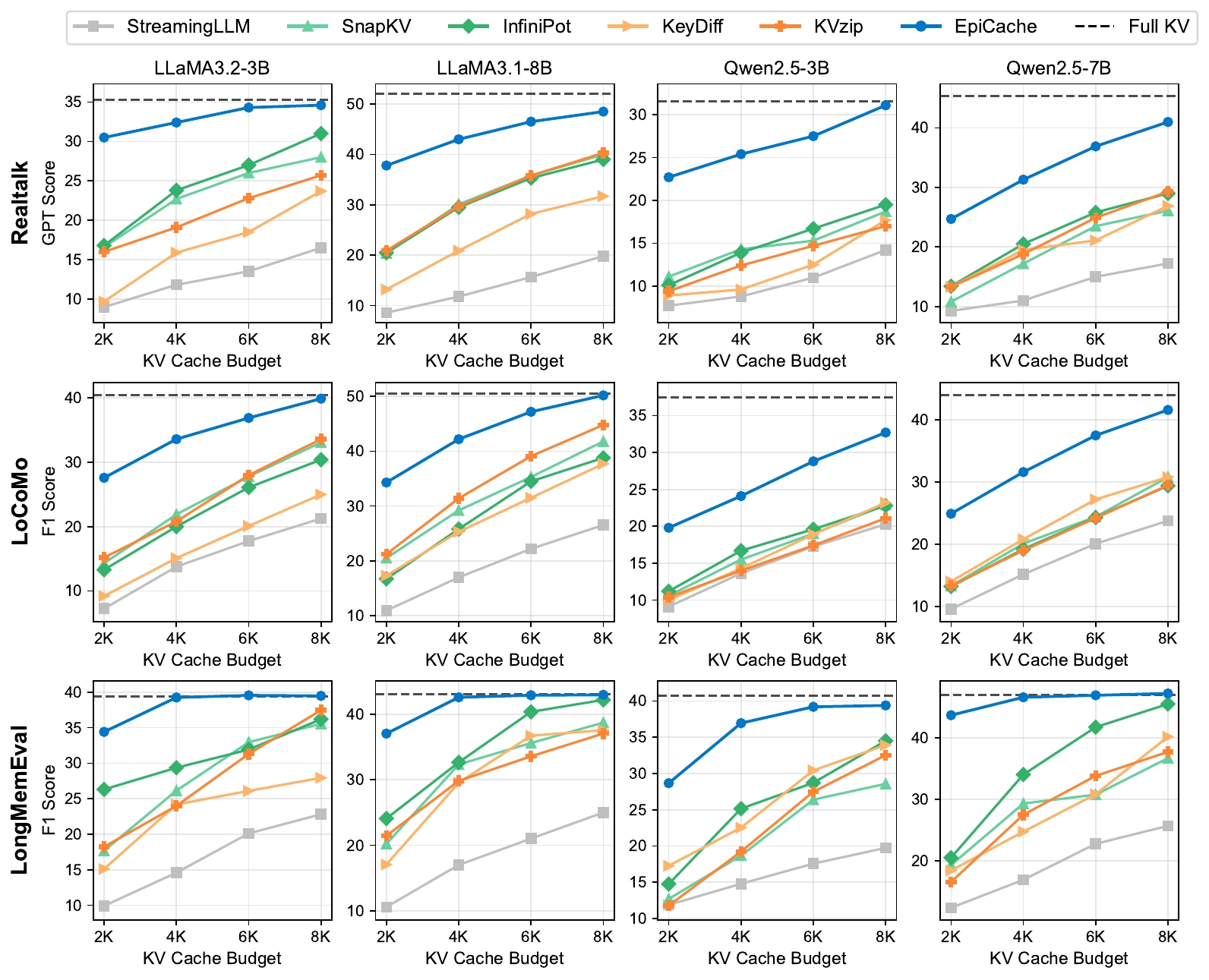}
    \caption{\textbf{Evaluation with Various Models} results with Realtalk, LoCoMo and LongMemEval. Evaluated with fixed KV cache budgets $M\in\{2\text{K},4\text{K},6\text{K},8\text{K}\}$ across four different models. The number of episodes (clusters) fixed to $E{=}4$ in all experiments.}
    \label{fig:appn_main_convqa}
    \vspace{-0.1in}
\end{figure}

%% file: Table/appn_longconvqa_llama.tex
\begin{table*}[t]
\centering
\resizebox{\textwidth}{!}
{
\begin{tabular}{ll|ccccc|cccc}
\toprule
 \textbf{LLaMA3.2-3B} & & \multicolumn{5}{c|}{\textbf{LoCoMo} (Full KV length: 21.8K)} & \multicolumn{4}{c}{\textbf{Realtalk} (Full KV length: 26.4K)} \\
\cmidrule(lr){3-7} \cmidrule(lr){8-11}
Method & $M$ & Multi-hop & Temporal & Open-domain & Single-hop & Avg & Multi-hop & Temporal & Common & Avg \\
\midrule
\multirow{1}{*}{Full KV} 
 & -- & 36.0 & 15.1 & 13.2 & 54.5 & 40.3 & 39.0 & 30.8 & 38.2 & 35.3 \\
\midrule
 & 2K & 17.3 & 3.7 & 17.3 & 17.6 & 14.3 & 20.1 & 10.5 & 25.0 & 16.6 \\
 SnapKV & 4K & 23.1 & 6.0 & 11.5 & 28.7 & 21.9 & 28.5 & 14.7 & 30.0 & 22.7 \\
 \citep{li2024snapkv} & 6K & 27.3 & 9.9 & 10.6 & 36.8 & 27.8 & 31.7 & 18.2 & 33.2 & 26.0 \\
 & 8K & 31.2 & 11.9 & 11.7 & 44.3 & 33.1 & 34.1 & 20.3 & 33.7 & 28.0 \\
\midrule
 & 2K & 15.9 & 7.2 & 10.1 & 15.0 & 13.3 & 22.1 & 9.4 & 24.3 & 16.8 \\
 InfiniPot & 4K & 21.3 & 12.0 & 10.0 & 23.7 & 20.0 & 29.2 & 16.0 & 31.9 & 23.8 \\
 \citep{kim-etal-2024-infinipot} & 6K & 25.6 & 17.0 & 11.9 & 31.3 & 26.1 & 32.8 & 19.7 & 32.4 & 27.0 \\
 & 8K & 28.6 & 20.5 & 11.4 & 36.9 & 30.4 & 35.6 & 24.5 & 36.9 & 31.0 \\
\midrule
 & 2K & 10.7 & 3.7 & 11.6 & 11.2 & 9.2 & 10.4 & 5.6 & 19.5 & 9.7 \\
 KeyDiff & 4K & 15.7 & 8.3 & 11.5 & 17.9 & 15.1 & 16.8 & 14.2 & 18.2 & 15.9 \\
 \citep{park2025keydiffkeysimilaritybasedkv} & 6K & 20.7 & 11.9 & 13.4 & 23.8 & 20.1 & 20.6 & 14.5 & 24.8 & 18.5 \\
 & 8K & 23.7 & 15.7 & 14.7 & 30.2 & 25.0 & 24.8 & 20.7 & 29.7 & 23.7 \\
\midrule
 & 2K & 22.0 & 4.5 & 12.0 & 17.3 & 15.2 & 18.7 & 9.5 & 27.9 & 16.0 \\
 KVzip & 4K & 21.9 & 13.3 & 10.0 & 24.5 & 20.8 & 25.0 & 10.1 & 29.4 & 19.1 \\
 \citep{kim2025kvzipqueryagnostickvcache} & 6K & 28.1 & 16.5 & 12.0 & 34.2 & 28.0 & 26.1 & 15.2 & 36.3 & 22.8 \\
 & 8K & 31.2 & 21.8 & 11.8 & 41.5 & 33.6 & 30.7 & 18.1 & 34.5 & 25.7 \\
\midrule
\rowcolor{}
\mscellf
\mscellf  & 2K & 29.3 & 15.9 & 13.7 & 33.1 & 27.6 & 34.7 & 23.0 & 41.3 & 30.5 \\ 
\mscells & 4K & 30.4 & 19.7 & 14.2 & 42.3 & 33.6 & 37.4 & 24.7 & 41.0 & 32.4 \\ 
\mscellt & 6K & 33.7 & 22.2 & 12.1 & 46.4 & 36.9 & 39.9 & 25.6 & 44.2 & 34.3 \\ 
\mscellfo \multirow{-4}{*}{\textbf{\methodname}} & 8K & 33.9 & 23.9 & 12.7 & 48.2 & 38.3 & 40.0 & 26.6 & 42.9 & 34.6 \\
\bottomrule \\
\end{tabular}
}

\resizebox{\textwidth}{!}{
\begin{tabular}{ll|ccccc|cccc}
\toprule
 \textbf{LLaMA3.1-8B} & & \multicolumn{5}{c|}{\textbf{LoCoMo} (Full KV length: 21.8K)} & \multicolumn{4}{c}{\textbf{Realtalk} (Full KV length: 26.4K)} \\
\cmidrule(lr){3-7} \cmidrule(lr){8-11}
Method & $M$ & Multi-hop & Temporal & Open-domain & Single-hop & Avg & Multi-hop & Temporal & Common & Avg \\
\midrule
\multirow{1}{*}{Full KV}
 & -- & 43.1 & 22.7 & 17.2 & 67.4 & 50.5 & 49.2 & 55.9 & 48.4 & 52.0 \\
\midrule
 & 2K & 23.4 & 6.6 & 14.3 & 25.5 & 20.5 & 20.2 & 15.8 & 34.3 & 20.4 \\
 SnapKV & 4K & 30.8 & 10.9 & 13.6 & 37.3 & 29.2 & 28.7 & 26.6 & 44.4 & 30.1 \\
 \citep{li2024snapkv} & 6K & 33.3 & 13.8 & 13.3 & 46.8 & 35.3 & 37.3 & 32.2 & 42.3 & 35.8 \\
 & 8K & 37.7 & 18.1 & 14.2 & 55.5 & 41.8 & 41.5 & 36.8 & 45.2 & 40.0 \\
\midrule
 & 2K & 15.2 & 12.3 & 10.2 & 19.6 & 16.7 & 21.2 & 14.9 & 35.0 & 20.5 \\
 InfiniPot & 4K & 23.0 & 22.1 & 13.9 & 29.5 & 25.8 & 30.0 & 25.8 & 39.1 & 29.5 \\
 \citep{kim-etal-2024-infinipot} & 6K & 29.8 & 29.6 & 13.9 & 40.1 & 34.5 & 34.4 & 33.3 & 43.4 & 35.3 \\
 & 8K & 33.1 & 33.8 & 13.8 & 45.5 & 38.8 & 38.3 & 38.2 & 43.4 & 39.0 \\
\midrule
 & 2K & 18.7 & 4.4 & 17.7 & 21.8 & 17.3 & 13.0 & 8.2 & 28.8 & 13.2 \\
 KeyDiff & 4K & 23.3 & 14.1 & 13.8 & 31.5 & 25.3 & 21.9 & 16.5 & 30.9 & 20.9 \\
 \citep{park2025keydiffkeysimilaritybasedkv} & 6K & 27.5 & 18.6 & 14.5 & 39.7 & 31.5 & 29.6 & 25.3 & 33.3 & 28.2 \\
 & 8K & 33.0 & 25.8 & 16.4 & 46.2 & 37.7 & 33.0 & 29.9 & 33.3 & 31.7 \\
\midrule
 & 2K & 24.4 & 8.4 & 22.8 & 24.7 & 21.2 & 20.6 & 16.6 & 34.6 & 20.9 \\
 KVzip & 4K & 28.8 & 24.3 & 15.3 & 36.8 & 31.4 & 29.5 & 26.4 & 40.0 & 29.7 \\
 \citep{kim2025kvzipqueryagnostickvcache} & 6K & 32.8 & 31.0 & 13.3 & 47.3 & 39.1 & 36.4 & 33.4 & 41.5 & 35.8 \\
 & 8K & 37.2 & 35.0 & 13.1 & 54.6 & 44.8 & 38.8 & 41.3 & 41.1 & 40.3 \\
\midrule
 \mscellf& 2K & 33.2 & 26.3 & 14.2 & 43.5 & 36.3 & 37.9 & 35.2 & 45.2 & 37.8 \\
 \mscells& 4K & 37.7 & 33.8 & 15.9 & 55.8 & 45.4 & 39.6 & 44.9 & 46.7 & 43.0 \\
 \mscellt& 6K & 38.2 & 36.4 & 16.4 & 58.2 & 47.3 & 42.3 & 50.5 & 35.2 & 46.5 \\
 \mscellfo \multirow{-4}{*}{\textbf{\methodname}} & 8K & 38.4 & 37.5 & 17.4 & 62.7 & 50.2 & 44.2 & 50.9 & 46.5 & 47.5 \\
\bottomrule
\end{tabular}
}
\vspace{.05in}
\caption{\textbf{LongConvQA (LoCoMo and Realtalk) Evaluation}: Comparison of different KV cache compression methods under block-prefill with LLaMA series models.}
\label{tab:appn_convqa_llama}
\end{table*}

%% file: Table/appn_longconvqa_qwen.tex
\begin{table*}[t]
\centering
\resizebox{\textwidth}{!}
{
\begin{tabular}{ll|ccccc|cccc}
\toprule
 \textbf{Qwen2.5-3B} & & \multicolumn{5}{c|}{\textbf{LoCoMo} (Full KV length: 21.9K)} & \multicolumn{4}{c}{\textbf{Realtalk} (Full KV length: 26.6K)} \\
\cmidrule(lr){3-7} \cmidrule(lr){8-11}
Method & $M$ & Multi-hop & Temporal & Open-domain & Single-hop & Avg & Multi-hop & Temporal & Common & Avg \\
\midrule
\multirow{1}{*}{Full KV} 
 & -- & 33.2 & 22.9 & 12.3 & 49.1 & 38.4 & 32.7 & 28.0 & 39.6 & 31.6 \\
\midrule
 & 2K & 14.1 & 8.0 & 11.5 & 10.4 & 10.6 & 12.4 & 5.6 & 23.5 & 11.1 \\
 SnapKV & 4K & 17.9 & 12.7 & 11.5 & 16.2 & 15.5 & 15.4 & 9.0 & 27.0 & 14.3 \\
 \citep{li2024snapkv} & 6K & 21.3 & 13.1 & 13.7 & 21.1 & 19.0 & 17.9 & 9.3 & 26.2 & 15.3 \\
 & 8K & 23.2 & 14.3 & 12.1 & 27.4 & 22.9 & 21.6 & 12.6 & 28.8 & 18.7 \\
\midrule 
 & 2K & 12.4 & 16.1 & 12.6 & 8.8 & 11.2 & 8.6 & 6.1 & 26.3 & 10.1 \\
 InfiniPot & 4K & 18.2 & 19.7 & 10.2 & 15.7 & 16.7 & 15.4 & 8.9 & 24.3 & 13.9 \\
 \citep{kim-etal-2024-infinipot} & 6K & 20.0 & 20.0 & 13.9 & 19.9 & 19.6 & 17.5 & 12.0 & 28.2 & 16.7 \\
 & 8K & 23.9 & 18.8 & 13.2 & 25.1 & 22.8 & 23.9 & 11.4 & 31.1 & 19.5 \\
\midrule
 & 2K & 10.4 & 14.2 & 13.6 & 7.8 & 10.0 & 3.9 & 9.0 & 23.5 & 8.9 \\
 KeyDiff & 4K & 14.0 & 15.8 & 12.0 & 14.1 & 14.3 & 9.2 & 6.0 & 21.6 & 9.6 \\
 \citep{park2025keydiffkeysimilaritybasedkv} & 6K & 19.6 & 16.4 & 9.2 & 20.8 & 18.9 & 11.5 & 10.0 & 22.6 & 12.5 \\
 & 8K & 21.6 & 17.3 & 7.4 & 27.8 & 23.2 & 18.1 & 15.1 & 24.4 & 17.7 \\
\midrule
 & 2K & 11.8 & 6.1 & 11.9 & 11.4 & 10.4 & 10.6 & 5.3 & 18.2 & 9.4 \\
 KVzip & 4K & 16.2 & 10.3 & 12.6 & 14.7 & 14.0 & 13.3 & 8.4 & 21.7 & 12.4 \\
 \citep{kim2025kvzipqueryagnostickvcache} & 6K & 19.5 & 12.7 & 11.3 & 19.1 & 17.4 & 15.0 & 10.0 & 27.8 & 14.7 \\
 & 8K & 21.9 & 14.3 & 14.1 & 24.3 & 21.1 & 18.8 & 11.4 & 28.4 & 17.0 \\
\midrule
 \mscellf & 2K & 23.6 & 7.6 & 13.4 & 23.8 & 19.8 & 25.6 & 13.2 & 37.7 & 22.0 \\
 \mscells & 4K & 27.1 & 10.7 & 11.4 & 29.7 & 24.1 & 30.0 & 17.1 & 36.9 & 25.4 \\
 \mscellt & 6K & 27.3 & 13.3 & 10.8 & 37.2 & 28.8 & 32.6 & 19.7 & 36.4 & 27.5 \\
 \mscellfo \multirow{-4}{*}{\textbf{\methodname}} & 8K & 31.3 & 17.0 & 10.0 & 41.7 & 32.7 & 33.9 & 25.0 & 41.1 & 31.1 \\
\bottomrule \\
\end{tabular}
}

\resizebox{\textwidth}{!}{
\begin{tabular}{ll|ccccc|cccc}
\toprule
 \textbf{Qwen2.5-7B} & & \multicolumn{5}{c|}{\textbf{LoCoMo} (Full KV length: 21.9K)} & \multicolumn{4}{c}{\textbf{Realtalk} (Full KV length: 26.6K)} \\
\cmidrule(lr){3-7} \cmidrule(lr){8-11}
Method & $M$ & Multi-hop & Temporal & Open-domain & Single-hop & Avg & Multi-hop & Temporal & Common & Avg \\
\midrule
\multirow{1}{*}{Full KV}
 & -- & 36.2 & 19.2 & 16.6 & 59.3 & 44.1 & 38.7 & 52.3 & 43.4 & 45.3 \\
\midrule
& 2K & 17.6 & 5.9 & 12.6 & 14.7 & 13.3 & 9.8 & 10.5 & 14.2 & 10.8 \\
SnapKV & 4K & 23.8 & 8.0 & 13.1 & 24.2 & 20.1 & 15.8 & 16.0 & 24.8 & 17.2 \\
\citep{li2024snapkv} & 6K & 27.8 & 9.1 & 15.2 & 30.2 & 24.4 & 20.6 & 24.4 & 29.2 & 23.5 \\
& 8K & 30.8 & 13.2 & 14.9 & 39.4 & 30.8 & 24.4 & 25.1 & 33.5 & 26.1 \\
\midrule
& 2K & 15.1 & 14.2 & 12.0 & 12.4 & 13.2 & 10.4 & 15.7 & 15.2 & 13.4 \\
InfiniPot & 4K & 19.2 & 20.0 & 10.8 & 19.9 & 19.2 & 17.1 & 22.7 & 23.7 & 20.5 \\
\citep{kim-etal-2024-infinipot} & 6K & 23.4 & 23.7 & 15.0 & 26.0 & 24.3 & 21.6 & 30.0 & 25.5 & 25.8 \\
& 8K & 27.7 & 14.0 & 14.0 & 32.7 & 29.4 & 26.3 & 29.4 & 31.5 & 29.0 \\
\midrule
& 2K & 12.8 & 16.7 & 12.4 & 13.6 & 14.0 & 9.8 & 15.0 & 18.4 & 13.3 \\
KeyDiff & 4K & 18.9 & 22.0 & 13.9 & 21.8 & 20.8 & 12.0 & 19.9 & 26.3 & 19.5 \\
\citep{park2025keydiffkeysimilaritybasedkv} & 6K & 26.3 & 25.2 & 14.3 & 29.8 & 27.2 & 15.7 & 26.1 & 21.1 & 21.1 \\
& 8K & 28.3 & 23.7 & 15.9 & 36.4 & 30.8 & 20.3 & 32.7 & 28.0 & 26.9 \\
\midrule
& 2K & 14.3 & 13.7 & 11.7 & 12.9 & 13.3 & 12.2 & 10.8 & 23.9 & 13.3 \\
KVzip & 4K & 19.4 & 16.2 & 13.9 & 20.6 & 19.0 & 17.7 & 16.0 & 29.8 & 18.8 \\
\citep{kim2025kvzipqueryagnostickvcache} & 6K & 24.3 & 19.5 & 12.5 & 27.2 & 24.2 & 22.5 & 24.6 & 32.6 & 24.9 \\
& 8K & 25.6 & 23.8 & 13.0 & 34.8 & 29.5 & 26.8 & 28.7 & 37.7 & 29.3 \\
\midrule
\mscellf & 2K & 26.4 & 15.4 & 13.2 & 29.3 & 24.9 & 24.1 & 21.3 & 36.6 & 24.7 \\
\mscells & 4K & 29.0 & 20.9 & 14.1 & 38.5 & 31.6 & 31.1 & 29.3 & 37.5 & 31.3 \\
\mscellt & 6K & 32.6 & 24.6 & 15.1 & 46.5 & 37.5 & 33.0 & 39.4 & 40.6 & 36.9 \\
\mscellfo \multirow{-4}{*}{\textbf{\methodname}} & 8K & 32.6 & 28.1 & 15.3 & 52.7 & 41.6 & 33.8 & 46.9 & 43.6 & 41.0 \\
\bottomrule
\end{tabular}
}
\vspace{.05in}
\caption{\textbf{LongConvQA (LoCoMo and Realtalk) Evaluation}: Comparison of different KV cache compression methods under block-prefill with Qwen series models.}
\label{tab:appn_convqa_qwen}
\end{table*}

%% file: Table/appn_longmemeval_llama_qwen.tex
\begin{table*}[t]
\centering
\resizebox{\textwidth}{!}{
\begin{tabular}{l c | *{8}{r} | *{8}{r}}
\toprule
\multirow{2}{*}{Method} & \multirow{2}{*}{$M$} 
& \multicolumn{8}{c|}{\textbf{LLaMA3.2-3B}} 
& \multicolumn{8}{c}{\textbf{LLaMA3.1-8B}} \\ 
& & SH & TH & MS & TR-E & TR-I & KU & IP & Avg.
  & SH & TH & MS & TR-E & TR-I & KU & IP & Avg. \\
\midrule
Full KV & 21K 
& 84.6 & 10.0 & 12.5 & 47.9 & 27.1 & 52.3 & 6.2 & 39.4
& 87.2 & 14.1 & 17.6 & 56.5 & 28.5 & 56.1 & 6.3 & 43.1 \\
\midrule
& 2K 
& 26.9 & 0.8 & 1.8 & 26.6 & 17.4 & 31.4 & 6.1 & 17.7
& 35.6 & 3.8 & 3.3 & 29.8 & 23.3 & 25.4 & 8.8 & 20.2 \\
SnapKV  & 4K 
& 40.0 & 5.3 & 2.4 & 40.2 & 20.4 & 48.5 & 6.5 & 26.1
& 63.3 & 9.7 & 3.9 & 45.8 & 24.4 & 45.8 & 10.8 & 32.4 \\
\citep{li2024snapkv} & 6K 
& 54.5 & 5.3 & 15.8 & 37.1 & 23.0 & 58.7 & 6.3 & 33.0
& 65.1 & 8.3 & 9.0 & 49.4 & 30.4 & 50.5 & 11.8 & 35.6 \\
& 8K 
& 67.1 & 7.9 & 12.4 & 37.1 & 27.1 & 56.6 & 7.1 & 35.6
& 74.3 & 13.1 & 13.6 & 53.8 & 25.9 & 52.5 & 10.9 & 38.7 \\
\midrule
& 2K 
& 46.2 & 0.8 & 10.8 & 40.5 & 19.1 & 40.7 & 8.9 & 26.3
& 39.8 & 1.8 & 3.4 & 29.1 & 31.7 & 35.8 & 7.6 & 24.1 \\
InfiniPot  & 4K 
& 48.5 & 7.9 & 12.3 & 33.6 & 18.4 & 52.3 & 7.6 & 29.4
& 62.4 & 7.9 & 4.1 & 53.9 & 23.0 & 46.8 & 9.2 & 32.7 \\
\citep{kim-etal-2024-infinipot} & 6K 
& 60.0 & 2.6 & 12.4 & 33.6 & 25.9 & 51.7 & 6.7 & 31.9
& 81.3 & 11.0 & 13.3 & 54.9 & 28.1 & 54.9 & 5.6 & 40.4 \\
& 8K 
& 76.0 & 4.4 & 13.3 & 40.7 & 25.0 & 52.9 & 7.6 & 36.2
& 90.3 & 9.1 & 18.8 & 54.0 & 28.5 & 56.8 & 9.4 & 42.2 \\
\midrule
& 2K 
& 35.3 & 0.5 & 4.1 & 34.3 & 17.7 & 5.7 & 2.6 & 15.1
& 26.0 & 6.8 & 9.5 & 28.8 & 12.3 & 22.8 & 5.5 & 17.1 \\
KeyDiff  & 4K 
& 54.2 & 2.1 & 2.4 & 34.3 & 15.4 & 34.4 & 6.8 & 24.2
& 60.0 & 14.7 & 12.9 & 41.4 & 20.2 & 31.8 & 6.1 & 29.6 \\
\citep{park2025keydiffkeysimilaritybasedkv} & 6K 
& 55.8 & 6.5 & 7.4 & 54.3 & 11.4 & 32.2 & 9.0 & 26.9
& 69.2 & 11.2 & 13.6 & 48.4 & 26.1 & 52.1 & 8.5 & 36.7 \\
& 8K 
& 56.1 & 2.1 & 6.6 & 37.1 & 25.9 & 37.7 & 7.7 & 27.9
& 70.7 & 11.2 & 18.4 & 46.1 & 30.0 & 50.1 & 5.2 & 37.6 \\
\midrule
& 2K 
& 30.8 & 0.0 & 1.8 & 30.9 & 15.2 & 30.3 & 7.6 & 18.2
& 33.8 & 7.5 & 8.0 & 36.0 & 19.4 & 27.2 & 9.1 & 21.5 \\
KVzip & 4K 
& 44.7 & 2.6 & 6.5 & 37.1 & 15.0 & 37.8 & 7.4 & 24.0
& 56.6 & 9.7 & 6.7 & 35.3 & 24.1 & 42.7 & 12.4 & 29.8 \\
\citep{kim2025kvzipqueryagnostickvcache} & 6K 
& 58.1 & 5.3 & 12.4 & 37.1 & 21.3 & 50.8 & 6.2 & 31.3
& 61.3 & 11.8 & 6.9 & 42.7 & 29.4 & 47.5 & 11.3 & 33.6 \\
& 8K 
& 73.5 & 7.9 & 12.5 & 40.7 & 26.2 & 59.1 & 6.9 & 37.5
& 73.6 & 14.4 & 7.7 & 48.6 & 26.9 & 51.4 & 6.2 & 37.1 \\
\midrule
\mscellf & 2K 
& 73.0 & 10.5 & 7.4 & 40.5 & 21.0 & 50.8 & 6.0 & 34.4
& 72.3 & 16.7 & 3.3 & 46.5 & 24.0 & 56.4 & 10.5 & 37.1 \\
\mscells & 4K 
& 79.9 & 12.6 & 16.6 & 41.4 & 27.0 & 53.9 & 9.1 & 39.3
& 87.2 & 14.1 & 17.4 & 56.5 & 28.5 & 54.6 & 3.9 & 42.6 \\
\mscells & 6K 
& 85.0 & 10.0 & 13.4 & 40.7 & 27.2 & 55.1 & 8.3 & 39.6
& 83.8 & 13.8 & 25.4 & 55.7 & 25.3 & 54.9 & 10.7 & 42.9 \\
\mscellfo \multirow{-4}{*}{\textbf{\methodname}} & 8K 
& 85.0 & 10.0 & 12.5 & 40.7 & 26.6 & 56.6 & 6.2 & 39.5
& 88.2 & 13.5 & 17.5 & 56.5 & 28.5 & 55.2 & 6.4 & 43.0 \\ \bottomrule
\\
\end{tabular}}

\resizebox{\textwidth}{!}{
\begin{tabular}{l c | *{8}{r} | *{8}{r}}
\toprule
\multirow{2}{*}{Method} & \multirow{2}{*}{$M$} 
& \multicolumn{8}{c|}{\textbf{Qwen2.5-3B}} 
& \multicolumn{8}{c}{\textbf{Qwen2.5-7B}} \\ 
& & SH & TH & MS & TR-E & TR-I & KU & IP & Avg.
  & SH & TH & MS & TR-E & TR-I & KU & IP & Avg. \\
\midrule
Full KV & 21K 
& 80.8 & 14.0 & 15.0 & 50.2 & 23.7 & 59.0 & 9.1 & 40.7
& 88.6 & 39.7 & 35.1 & 32.9 & 35.4 & 47.9 & 12.9 & 46.9 \\
\midrule
& 2K 
& 23.5 & 1.2 & 1.3 & 11.9 & 0.1 & 25.6 & 3.8 & 12.7
& 19.9 & 4.9 & 8.3 & 33.1 & 21.5 & 31.1 & 8.9 & 19.3 \\
SnapKV & 4K 
& 48.8 & 6.5 & 0.8 & 11.9 & 1.1 & 27.9 & 4.6 & 18.7
& 42.9 & 10.9 & 22.0 & 33.6 & 28.8 & 38.4 & 10.7 & 29.3 \\
\citep{li2024snapkv} & 6K 
& 59.3 & 3.9 & 19.6 & 32.4 & 5.9 & 34.5 & 5.7 & 26.4
& 50.4 & 10.9 & 17.0 & 32.9 & 28.4 & 42.6 & 11.3 & 30.7 \\
& 8K 
& 56.3 & 11.4 & 20.4 & 29.8 & 7.9 & 41.7 & 8.1 & 28.6
& 67.4 & 17.1 & 24.4 & 32.9 & 31.6 & 42.9 & 14.1 & 36.7 \\
\midrule
& 2K 
& 31.0 & 1.2 & 11.9 & 22.6 & 5.4 & 17.5 & 4.9 & 14.8
& 29.8 & 1.6 & 2.0 & 33.3 & 23.1 & 32.2 & 11.8 & 20.5 \\
InfiniPot & 4K 
& 46.2 & 5.3 & 16.5 & 33.3 & 19.4 & 31.5 & 6.8 & 25.2
& 44.7 & 22.6 & 25.3 & 36.4 & 36.7 & 41.6 & 10.3 & 34.0 \\
\citep{kim-etal-2024-infinipot} & 6K 
& 55.9 & 4.8 & 10.6 & 32.4 & 25.7 & 40.1 & 6.1 & 28.7
& 66.1 & 26.7 & 32.9 & 36.4 & 40.5 & 48.1 & 11.7 & 41.7 \\
& 8K 
& 70.2 & 6.1 & 15.7 & 35.9 & 27.3 & 47.3 & 7.7 & 34.5
& 78.6 & 34.0 & 33.6 & 32.9 & 41.4 & 50.1 & 13.1 & 45.5 \\
\midrule
& 2K 
& 12.6 & 14.0 & 6.3 & 44.1 & 11.1 & 27.6 & 3.3 & 17.3
& 18.1 & 8.2 & 9.4 & 31.1 & 14.3 & 31.5 & 11.7 & 18.4 \\
KeyDiff & 4K 
& 30.5 & 11.4 & 11.1 & 34.5 & 18.9 & 33.0 & 5.7 & 22.5
& 43.1 & 5.8 & 9.9 & 37.1 & 20.5 & 32.7 & 10.7 & 24.7 \\
\citep{park2025keydiffkeysimilaritybasedkv} & 6K 
& 51.6 & 15.3 & 21.1 & 43.7 & 20.5 & 35.4 & 9.6 & 30.4
& 57.5 & 13.0 & 13.9 & 34.2 & 26.8 & 38.0 & 9.9 & 30.8 \\
& 8K 
& 62.6 & 22.6 & 15.4 & 30.7 & 22.0 & 47.5 & 7.8 & 33.9
& 69.6 & 23.6 & 32.9 & 39.1 & 25.6 & 50.2 & 11.9 & 40.2 \\
\midrule
& 2K 
& 18.4 & 1.2 & 1.3 & 13.7 & 10.9 & 22.5 & 4.2 & 11.8
& 21.1 & 4.8 & 2.4 & 22.8 & 19.8 & 27.1 & 8.9 & 16.5 \\
KVzip & 4K 
& 39.4 & 6.5 & 5.0 & 23.8 & 13.8 & 25.3 & 6.4 & 19.2
& 45.4 & 14.1 & 22.0 & 31.7 & 23.3 & 29.9 & 11.5 & 27.5 \\
\citep{kim2025kvzipqueryagnostickvcache} & 6K 
& 51.9 & 1.2 & 20.0 & 44.1 & 17.6 & 34.3 & 5.7 & 27.5
& 57.8 & 23.5 & 22.8 & 33.6 & 31.9 & 34.4 & 11.2 & 33.8 \\
& 8K 
& 68.4 & 8.8 & 15.0 & 39.5 & 20.7 & 40.9 & 10.0 & 32.5
& 69.6 & 23.5 & 23.4 & 32.9 & 35.0 & 40.3 & 10.6 & 37.7 \\
\midrule
\mscellf & 2K 
& 52.6 & 3.5 & 10.0 & 32.4 & 22.9 & 46.7 & 6.7 & 28.7
& 70.4 & 36.9 & 31.1 & 42.4 & 35.9 & 48.5 & 12.4 & 43.7 \\
\mscells & 4K 
& 74.4 & 14.0 & 11.0 & 46.7 & 17.6 & 55.9 & 9.1 & 37.0
& 83.6 & 38.3 & 29.8 & 37.6 & 49.6 & 41.3 & 12.8 & 46.6 \\
\mscellt & 6K 
& 77.3 & 16.7 & 15.4 & 46.7 & 22.0 & 55.3 & 10.5 & 39.2
& 86.1 & 38.3 & 33.1 & 32.9 & 39.9 & 48.8 & 12.7 & 46.9 \\
\mscellfo \multirow{-4}{*}{\textbf{\methodname}} & 8K 
& 77.3 & 16.7 & 15.0 & 46.7 & 22.8 & 55.9 & 10.1 & 39.4
& 86.0 & 38.3 & 33.1 & 37.6 & 40.5 & 47.3 & 12.9 & 47.2 \\
\bottomrule
\end{tabular}}
\vspace{.05in}
\caption{\textbf{LongConvQA (LongMemEval) Evaluation}: Evaluation results with Qwen and LLaMA series models under block prefill. SH = Single Hop, TH = Two Hop, MS = Multi-Session, TR-E = Temporal Reasoning (explicit), TR-I = Temporal Reasoning (implicit), KU = Knowledge Update. IP = Implicit Preference.}
\label{tab:lme_side_by_side}
\end{table*}